\documentclass{article}
\usepackage{iclr2027_conference,times}

\usepackage{amsmath,amsfonts,bm}

\def\eqref#1{equation~\ref{#1}}

\def\1{\bm{1}}

\DeclareMathAlphabet{\mathsfit}{\encodingdefault}{\sfdefault}{m}{sl}
\SetMathAlphabet{\mathsfit}{bold}{\encodingdefault}{\sfdefault}{bx}{n}

\usepackage{url}
\usepackage{booktabs}
\usepackage{multirow}
\usepackage{graphicx}
\usepackage{xspace}
\usepackage{amsmath}
\usepackage{amssymb}
\usepackage{algorithm}
\usepackage{algpseudocode}
\usepackage{placeins}
\usepackage{CJKutf8}
\usepackage{marvosym}
\usepackage{hyperref}
\hypersetup{hidelinks}

\newcommand{\Rtgt}{R_{\mathrm{tgt}}}

\newcommand{\Ractual}{R_{\mathrm{stored}}}
\newcommand{\Bactual}{B_{\mathrm{stored}}}

\title{EntroPack: Fast and Accurate Entropy-Coded\\
Weight Compression at Arbitrary Bitrates}
\author{Hong Zhang \qquad Zhongjie Duan \qquad Yingda Chen\\
{\normalfont\small\texttt{\{hongzhang99,duanzhongjie.dzj,yingda.chen\}@alibaba-inc.com}}}
\iclrfinalcopy

\begin{document}
\maketitle
\lhead{Preprint}

\begin{abstract}
Weight compression helps large neural networks fit deployment memory
budgets, but common fixed-width formats offer only coarse storage choices.
Entropy coding supports finer rates, yet the achieved size depends on the
quantized weight distribution and coding overhead. Exploiting this
flexibility requires accurate rate selection and efficient weight
reconstruction for inference.
We present EntroPack, an entropy-coded weight compressor that supports
arbitrary target bitrates without activation calibration or fine-tuning.
It combines row-normalized $E_8$ lattice quantization with a conditional
probability model of lattice coordinates. Sampled storage estimates select
the quantization resolution without repeated full-stream encoding. The final
coordinates are entropy-coded in independently decodable tiles, enabling
fast, fused symbol decoding and numerical weight reconstruction on the GPU.
EntroPack supports floating-point and integer weight containers, such as
BF16, FP16, FP8, and INT8, with storage bitrate controlled independently
of numerical precision. Online decoding adds latency that grows with weight
count, making the method well suited to
compute-intensive workloads such as diffusion denoising and Transformer prefill. Experiments
demonstrate fast encoding and modest inference overhead in these settings.
When compressing the linear-layer weights of the image generator Z-Image-Turbo, EntroPack achieves substantially lower weight error and higher image fidelity relative to the BF16 reference than fixed-width formats at comparable storage rates, with modest denoising-step overhead. Targeting 4 bits per parameter, it achieves about 24\% lower relative $L_2$ weight error than NF4 while using less storage.
Source code is available at \url{https://github.com/modelscope/entropack}.
\end{abstract}

\section{Introduction}
\label{sec:intro}

Weight storage constrains the deployment of large neural networks.
Fixed-width formats, including INT8, FP8, and four-bit formats,
provide established compute paths, but a
deployment budget may lie between their available storage sizes
\citep{llmint8,qlora,fp8_formats,mx_formats}. Entropy coding allows finer
choices by assigning shorter codes to frequent quantized values: adjusting
quantization resolution changes both weight error and average storage.
This approach is well established in weight compression
\citep{han2015deep,deepcabac}, with recent methods supporting fractional rates
without activation calibration \citep{hrtn,hyperquant}.

Using this flexibility introduces two challenges. First, a requested
bitrate must be translated into a quantization resolution \citep{hrtn}, even though the
stored size also depends on the tensor's symbol distribution and coding
metadata. Repeatedly encoding full tensors during this search can
make compression expensive. Second, inference requires the compressed symbols
to be decoded and converted into numerical weights before matrix products.
The stream layout and weight representation must support parallel reconstruction
to limit added latency \citep{dfloat11}. A practical compressor needs
inexpensive storage estimation and efficient weight recovery.

In this paper, we present EntroPack, a calibration-free weight compressor
that combines fine-grained bitrate control with efficient GPU reconstruction.
After row normalization, it quantizes blocks of eight weights onto the $E_8$
lattice and represents the resulting points as invertible integer fields
\citep{conway_sloane_fast}. A conditional probability model captures
distribution differences between the lattice's integer and half-integer
cosets. Using this model and coding metadata, EntroPack
estimates storage from sampled rows and searches for a quantization scale
matching the requested bitrate, avoiding repeated full-stream encoding.
After scale selection, we apply row-scale fitting and optional per-row
rate--distortion refinement to further improve weight reconstruction.

\begin{figure}[!tbp]
\centering
\includegraphics[width=0.95\linewidth]{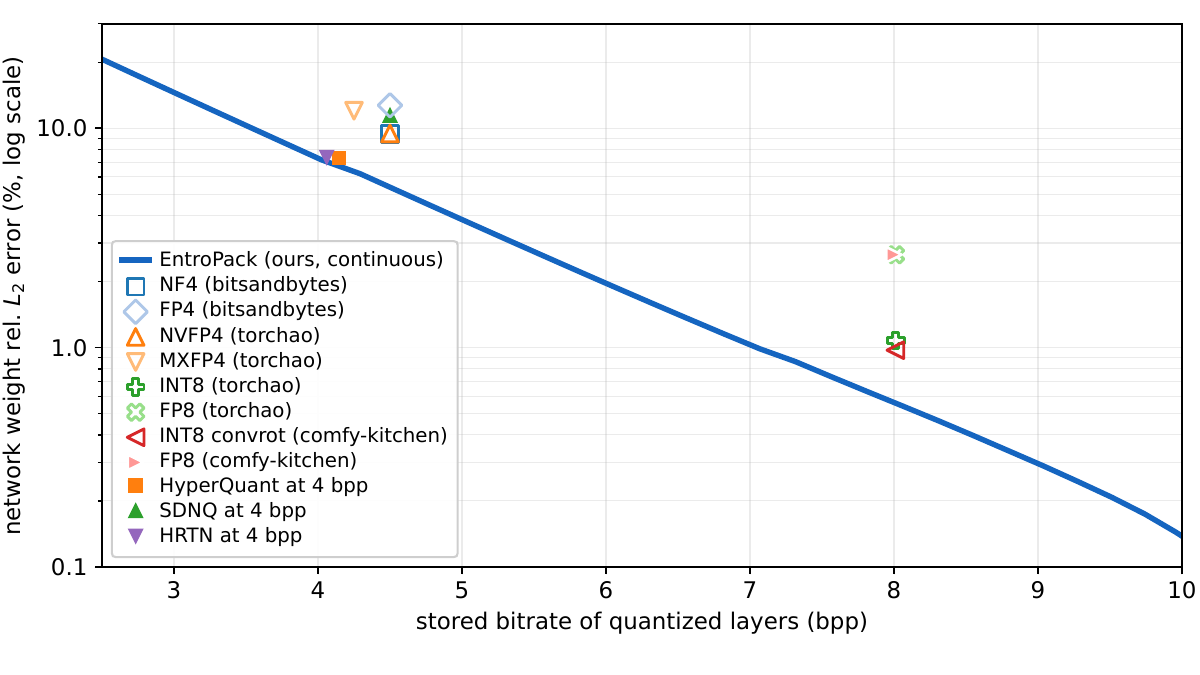}
\caption{Storage and reconstruction error of compressed linear weights in
Z-Image-Turbo.}
\label{fig:rd}
\end{figure}

EntroPack then encodes the fields with range asymmetric numeral systems
(rANS) in independently decodable tiles. This layout permits parallel
decoding across tiles while preserving the field order required by the
conditional model. The arithmetic inverse of the field representation allows
symbol decoding, lattice reconstruction, and row rescaling to be fused on
the GPU, without an intermediate symbol tensor or reconstruction codebook.
The decoder outputs weights for matrix products in floating-point
or integer formats, including FP8 and INT8 for low-precision computation.
Online decoding adds latency that grows with weight count, making EntroPack
well suited to compute-intensive workloads such as diffusion denoising and
Transformer prefill.
We evaluate weight reconstruction, inference quality, and runtime after
compressing linear weights of the image generator Z-Image-Turbo, the
audio-video generator MiniMax-H3, and the language model Qwen3.8-27B.
EntroPack supports continuously adjustable target bitrates, and
Figure~\ref{fig:rd} shows that it reconstructs Z-Image-Turbo's linear weights
with lower error than the tested fixed-width formats at comparable storage.
Our contributions are as follows:
\begin{itemize}
  \item \textbf{Fine-grained rate control.}
  EntroPack selects lattice quantization resolution from sampled storage
  estimates specific to each tensor. This provides user-specified weight
  bitrates without activation calibration or repeated full-stream encoding
  during rate selection.
  \item \textbf{Compact coding and GPU reconstruction.}
  Coset-conditioned coordinate coding and independent rANS tiles allow
  symbol decoding, arithmetic reconstruction, and row rescaling to be fused
  on the GPU. The representation requires neither an intermediate symbol
  tensor nor a reconstruction codebook and supports floating-point and
  integer outputs.
  \item \textbf{Compression quality and runtime.}
  EntroPack achieves lower weight error than
  tested fixed-width formats at comparable storage and faster conversion
  than tested entropy-coded methods. For Z-Image-Turbo's linear weights at a
  4 bpp target, conversion takes 1.12~s and reconstruction adds 7.8\% to denoising-step time
  relative to the uncompressed model.
\end{itemize}

\section{Related work}
\label{sec:related}

\paragraph{Entropy-coded quantization.}
Entropy-constrained quantization jointly considers coding rate and distortion
\citep{chou_ecvq,gray_neuhoff}. In weight compression, Deep Compression applies Huffman coding to
quantized weights \citep{han2015deep}, while DeepCABAC uses context-adaptive
arithmetic coding with rate--distortion optimization \citep{deepcabac}.
More recently, EntQuant reduces weight entropy before coding with asymmetric numeral systems
\citep{entquant}, and NeuZip
combines exponent coding with a precision-reduction option \citep{neuzip}.
For rate control, HRTN selects scalar quantization scales using Huffman code length
\citep{hrtn}. HyperQuant combines rotated lattices, algebraic constraint
removal, and Rice coding with a precomputed relationship between rate and
signal-to-noise ratio
\citep{hyperquant}.

\paragraph{Structured weight quantization.}
Weight quantizers differ in both their use of calibration data and their
representation of weights. GPTQ and AWQ use activation information for
low-bit quantization \citep{gptq,awq}. In terms of representation, NF4 provides a nonuniform
four-bit scalar format \citep{qlora}.
Beyond scalar formats, QuIP and QuIP\# combine weight transformations with structured quantization.
QuIP\# uses a shared $E_8$-based codebook to quantize blocks of eight
weights jointly \citep{quip,quipsharp}.
Other structured representations include additive codebooks, grouped and
Leech lattices, and trellis quantization
\citep{aqlm,grouped_lattice,leech_lattice,qtip}.

\paragraph{Deployment and GPU compression.}
Deployment-oriented systems such as SDNQ, Quanto, and GGUF provide quantized
storage and execution paths \citep{sdnq,quanto,gguf}.
At the storage level, lossless weight compressors exploit redundancy in
numerical representations to reduce storage without altering weight values
\citep{dfloat11,zipnn,shannonbound}.
For lossy compression, ZFP and cuSZp address precision- or error-controlled
compression of numerical arrays
\citep{zfp,cuszp}. For parallel entropy coding, DietGPU provides GPU rANS
primitives for symbol encoding and decoding \citep{dietgpu}, based on asymmetric numeral systems
\citep{duda_ans}.

\section{Method}
\label{sec:method}

\subsection{Objective and pipeline}
\label{sec:problem}

Let $W\in\mathbb{R}^{R\times C}$ be a weight matrix with $N=RC$ elements
stored in numerical data type (dtype) $\tau$, where $C$ is divisible by eight. EntroPack produces a compressed
representation $\mathcal{C}$ and reconstructs $\hat W$ in the same dtype.
We measure relative reconstruction error and stored rate as
\begin{equation}
\label{eq:objective}
 E_{\mathrm{rel}}(W,\hat W)
 =\sqrt{\frac{\sum_{r=1}^{R}\sum_{j=1}^{C}(\hat W_{rj}-W_{rj})^2}
 {\sum_{r=1}^{R}\sum_{j=1}^{C}W_{rj}^2}},
 \qquad
 \Ractual(\mathcal{C})=\frac{8\Bactual(\mathcal{C})}{N},
\end{equation}
where $\Bactual$ is the actual stored size in bytes. For a requested rate
$\Rtgt$ in bpp, the goal is to minimize reconstruction error while keeping
the stored rate close to the request. The dtype specifies numerical values independently of compressed bit width.

\begin{figure}[!htbp]
\centering
\includegraphics[width=0.98\linewidth]{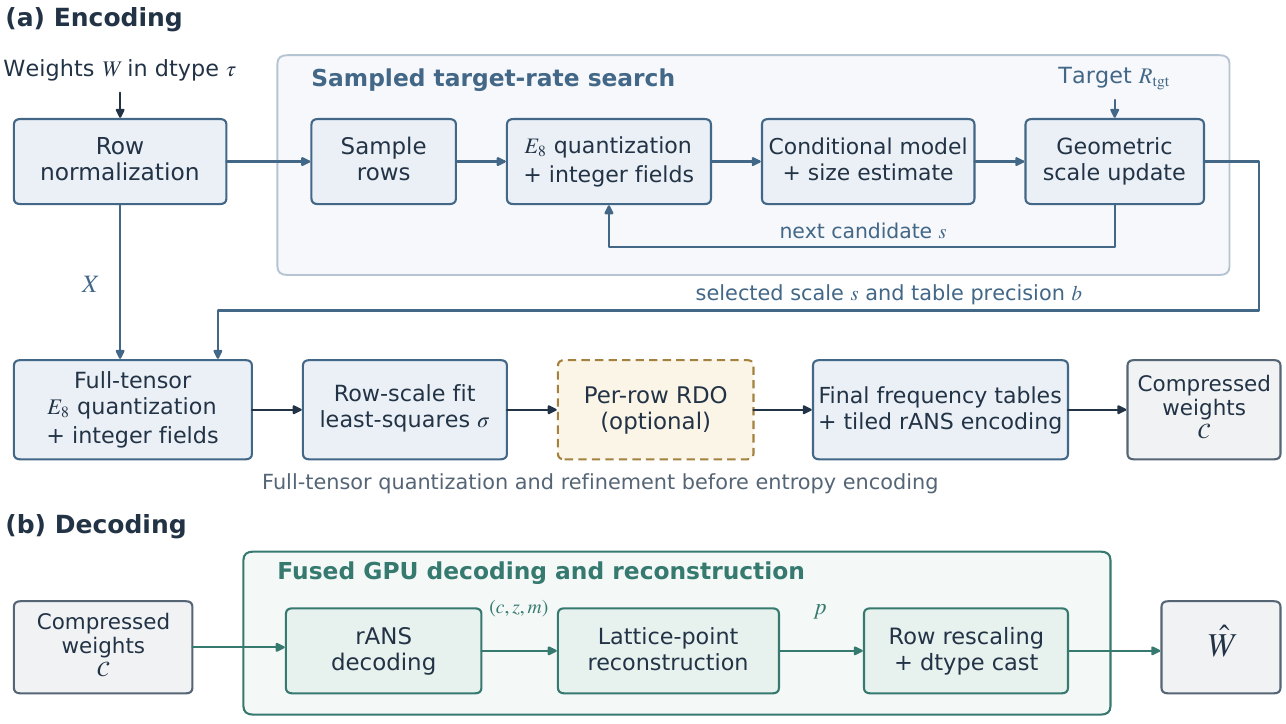}
\caption{EntroPack encoding and decoding. (a) Sampled scale selection precedes
full-tensor quantization, row-scale fitting, optional RDO, and tiled rANS
encoding. (b) Symbol decoding, lattice reconstruction, row rescaling, and
dtype conversion are fused on the GPU.}
\label{fig:pipeline}
\end{figure}

Figure~\ref{fig:pipeline} shows the compression and reconstruction pipeline.
Encoding starts by selecting a quantization scale on a sample of rows.
For each candidate scale, we map normalized weights to lattice points and
represent these points as integer fields, whose statistics determine an
estimated storage cost. At the selected scale, we quantize the full tensor
and fit one reconstruction scale per row. Optional rate--distortion
optimization (RDO) refines the per-row quantization before the final fields
are rANS-encoded into independent tiles. The resulting compressed
representation contains the tile payloads and the information needed to
decode them.

The decoding path reverses these operations: rANS decoding recovers the
fields, the algebraic inverse reconstructs lattice points, and row rescaling
and dtype conversion produce numerical weights.

\subsection{Row-normalized lattice representation}
\label{sec:lattice}

\paragraph{Normalization and nearest points.}
We normalize row magnitudes so that a common scale can control quantization
resolution across rows. The $E_8$ lattice combines the geometric efficiency
of vector quantization with a structured nearest-point algorithm
\citep{conway_sloane_book}.
We convert $W$ to FP32, denoted by $x$, and compute the
root-mean-square (RMS) magnitude of each row,
$\mu_r=\max(\sqrt{C^{-1}\sum_j x_{rj}^2},\eta)$, with $\eta=10^{-12}$.
We divide row $r$ by $\mu_r$ and partition it into vectors
$X_v\in\mathbb{R}^{8}$. A shared quantization scale $s>0$ determines their
lattice points:
\begin{equation}
\label{eq:quantize}
 p_v(s)=\operatorname{nearest}_{E_8}(X_v/s).
\end{equation}
For a block in row $r$, the corresponding weight approximation is
$s\mu_r p_v(s)$. Thus $s$ controls quantization in normalized coordinates,
while $\mu_r$ restores the row's original magnitude.

Write $D_8=\{z\in\mathbb{Z}^8:\sum_i z_i\equiv0\pmod 2\}$, so that
$E_8=D_8\cup(D_8+\tfrac12\mathbf{1})$. The nearest point in each coset is obtained
by coordinate rounding followed, when needed, by a parity correction
\citep{conway_sloane_fast}. For $Y=X_v/s$, let $f=\operatorname{round}(Y)$,
$d=Y-f$, $\pi=(\sum_i f_i)\bmod2$, and
$j^\star=\arg\max_j|d_j|$. With $e_j$ denoting the $j$th coordinate basis
vector,
\begin{equation}
\label{eq:d8}
 \operatorname{nearest}_{D_8}(Y)
 =f+\pi\operatorname{sign}(d_{j^\star})e_{j^\star}.
\end{equation}
We compare this candidate with
$\operatorname{nearest}_{D_8}(Y-\tfrac12\mathbf{1})+\tfrac12\mathbf{1}$ and select
the closer one. Coordinate ties round to the nearest even integer.
Parity correction breaks equal-residual ties by the first coordinate and
uses $\operatorname{sign}(0)=+1$. These conventions fix the point recovered
during decoding.

\paragraph{An invertible field representation.}
\label{sec:source}
Each $p_v$ lies in one of the two cosets. Let $c\in\{0,1\}$ identify that coset and
write $z=p_v-\tfrac c2\mathbf{1}\in D_8$. The eighth integer coordinate satisfies
a parity constraint determined by the first seven. Define
\begin{equation}
\label{eq:parity}
 \textstyle \pi=(\sum_{i=1}^{7}z_i)\bmod2,\qquad m=(z_8-\pi)/2.
\end{equation}
The point is represented by $(c,z_1,\ldots,z_7,m)$ and recovered as
\begin{equation}
\label{eq:point_inverse}
 p_v=(z_1,\ldots,z_7,2m+\pi)+\tfrac c2\mathbf{1}.
\end{equation}
The fields uniquely identify each point and permit arithmetic recovery.
In EntroPack, this representation supplies the symbols for the conditional
entropy model described next.

\subsection{Coset-conditioned probability model}
\label{sec:entropy}

The coordinate fields can have different distributions in the two cosets.
We capture this dependence by conditioning their probabilities on the coset.
Let $a=(z_1,\ldots,z_7,m)$ denote the eight coordinate fields, and let
$\theta$ collect the distribution parameters. We model the coset and
coordinates as
\begin{equation}
\label{eq:cond}
 q_\theta(c,a)=q_\theta(c)\prod_{i=1}^{8}q_\theta(a_i\mid c),
\end{equation}
using seventeen categorical distributions: one for the coset and two for
each coordinate field. For any given set of quantized vectors, we estimate
these distributions from the corresponding field counts.

The pooled alternative codes the coset and each coordinate field independently.
With exact empirical probabilities, conditioning reduces its ideal coding cost by
\begin{equation}
\label{eq:conditioning_gain}
 \textstyle [H(c)+\sum_i H(a_i)]-[H(c)+\sum_i H(a_i\mid c)]
 =\sum_i I(a_i;c),
\end{equation}
where $H$ denotes entropy and $I$ mutual information, both in bits. The model
captures dependence between each coordinate field and the coset, while
treating the coordinates as conditionally independent. The net stored-rate
gain also depends on frequency quantization and table storage.

We represent probabilities by normalizing counts to integer frequencies
summing to $2^b$, where $b$ is the probability-table
precision in bits.
Every observed symbol receives a positive frequency. Let $n_{t,z}(s)$ be
the count of symbol $z$ in distribution $t$ at quantization scale $s$, and
let $f_{t,z}(s)$ be its normalized frequency. The probability
is $f_{t,z}(s)/2^b$, giving the modeled code length in bits
\begin{equation}
\label{eq:length}
 C_{\mathrm{CE}}(s,b)
 =\sum_t\sum_{z:n_{t,z}>0}
 n_{t,z}(s)\bigl(b-\log_2 f_{t,z}(s)\bigr).
\end{equation}
This quantity is evaluated from symbol counts and provides the coding-cost
term for scale selection.

\subsection{Target-rate scale selection}
\label{sec:optimizer}

\paragraph{Scale--rate relationship.}
The scale $s$ changes both the lattice resolution and the distribution of coded
fields. Under a smooth continuous-source approximation and sufficiently fine
quantization, the ideal rate of an eight-dimensional scaled lattice is
\begin{equation}
\label{eq:high_resolution}
 R_{\mathrm{ideal}}(s)\approx h(X)/8-\log_2s-(\log_2v_\Lambda)/8.
\end{equation}
Here $h(X)$ is the source differential entropy in bits and $v_\Lambda$ the
lattice cell volume, with $v_{E_8}=1$ \citep{gray_neuhoff}. This motivates
geometric scale search.

\paragraph{Stored-size estimate.}
Stored size includes coded symbols and the metadata needed for
reconstruction. We estimate it from candidate field counts and the planned
tile layout, without producing a codestream.
The model cost $C_{\mathrm{CE}}$ estimates the symbol cost,
but some of this information remains in the coder's final states rather
than in emitted payload words. We estimate the payload after this correction,
then add the stored states and other metadata. In the tiled rANS format
(\S\ref{sec:rans-encode}), each independently decodable tile uses $J$ coder
states and 16-bit payload words. For $n_{\mathrm{tile}}$ tiles,
\begin{equation}
\label{eq:estimate}
 \widehat L_{\mathrm{payload}}
 =16\lceil\max(0,C_{\mathrm{CE}}-16J n_{\mathrm{tile}})/16\rceil.
\end{equation}
The subtraction approximates information retained in those states, and
the remainder is rounded to whole payload words. Adding auxiliary storage gives
\begin{equation}
\label{eq:outer_estimate}
 \widehat L=
 \widehat L_{\mathrm{payload}}+L_{\mathrm{restart}}
 +L_{\mathrm{tables}}+L_{\mathrm{scales}}+\widehat L_{\mathrm{other}},
 \qquad \widehat B=\widehat L/8.
\end{equation}
Here $L_{\mathrm{restart}}$ covers the final states and payload offsets
needed to start decoding each tile. The remaining terms account for
probability tables, row scales, and other metadata.

\paragraph{Sampled rate evaluation.}
We reuse a fixed sample of complete rows across candidate scales. At each
$(s,b)$, we quantize the sample, obtain $C_{\mathrm{CE}}(s,b)$ from field
counts, and apply the stored-size estimate above. For
$N_{\mathrm{sub}}$ sampled weights and estimated size
$\widehat B_{\mathrm{sub}}(s,b)$, scale selection seeks
\begin{equation}
\label{eq:rootfind}
 \widehat R_{\mathrm{sub}}(s,b)
 =8\widehat B_{\mathrm{sub}}(s,b)/N_{\mathrm{sub}}
 \approx\Rtgt.
\end{equation}
This avoids encoding full streams for candidate scales.
Appendix~\ref{app:search} specifies the sampling settings.

\paragraph{Geometric search.}
At a fixed probability precision $b$, we evaluate the geometric midpoint
$s=\sqrt{lo\,hi}$ of a scale bracket. When the estimated rate exceeds the
target, we replace $lo$ by $s$. Otherwise, we replace $hi$ by $s$. After a
fixed number of iterations, we use the geometric midpoint of the remaining bracket
to quantize the full tensor. If its field alphabets exceed the capacity of
the probability representation, we increase $b$ and repeat the search,
coarsening $s$ if the maximum supported precision is reached.
Algorithm~\ref{alg:rate} summarizes scale selection and
full-tensor quantization. $\mathcal B$ lists probability precisions in
ascending order and $I_s$ is the number of scale-search steps.

\begin{algorithm}[!htbp]
\caption{Target-rate scale selection and full-tensor quantization.}
\label{alg:rate}
\begin{algorithmic}[1]
\Require Tensor $W$, target $\Rtgt$, precisions $\mathcal B$, search steps $I_s$
\State Normalize rows to obtain $X$ and fix a row sample $\mathcal S$ of $N_{\mathrm{sub}}$ weights
\For{$b\in\mathcal B$ in ascending order}
  \State Initialize the scale bracket $(lo,hi)$
  \For{$i=1,\ldots,I_s$}
    \State $s\gets\sqrt{lo\,hi}$
    \State Quantize $X_{\mathcal S}/s$ and obtain field counts $n_{t,z}$ and $b$-bit frequencies $f_{t,z}$
    \State $C_{\mathrm{CE}}\gets\sum_{t,z:n_{t,z}>0}n_{t,z}\bigl(b-\log_2f_{t,z}\bigr)$
    \State Estimate stored bytes $\widehat B_{\mathrm{sub}}(s,b)$ from $C_{\mathrm{CE}}$ and format overheads
    \State $\widehat R_{\mathrm{sub}}\gets8\widehat B_{\mathrm{sub}}(s,b)/N_{\mathrm{sub}}$
    \State Set $lo\gets s$ if $\widehat R_{\mathrm{sub}}>\Rtgt$ and $hi\gets s$ otherwise
  \EndFor
  \State Set $s\gets\sqrt{lo\,hi}$ and quantize all blocks to form their fields
  \State Stop the precision search when the full-tensor alphabets fit $2^b$
\EndFor
\State \Return $s$, $b$, lattice points $\{p_v\}$, and fields $\{(c_v,a_v)\}$
\end{algorithmic}
\end{algorithm}

\FloatBarrier
\subsection{Row-scale refit and rate--distortion optimization}
\label{sec:rdo}
\label{sec:scales}

\paragraph{Row-scale refit.}
The search scale $s$ determines which lattice points are selected. Once those
points are fixed, a separate reconstruction scale can reduce row error
without changing their coded fields. Writing $p_{rj}$ for the lattice
coordinates of row $r$, we fit this scale by least squares:
\begin{equation}
\label{eq:scale_fit}
 \sigma_r=
 \begin{cases}
 \max\!\left(\eta,\dfrac{\sum_j x_{rj}p_{rj}}{\sum_j p_{rj}^2}\right),
     & \sum_j p_{rj}^2>0,\\
 s\mu_r,&\text{otherwise}.
 \end{cases}
\end{equation}
For nonzero lattice rows, this minimizes
$\sum_j(x_{rj}-\sigma_r p_{rj})^2$ over scales at least $\eta$.
The fitted $\sigma_r$ replaces $s\mu_r$, restoring the row's numerical
scale without changing its coded symbols or scale storage.

\paragraph{Per-row rate--distortion refinement.}
Rows can have different rate--distortion trade-offs even after normalization.
RDO compares several quantization resolutions per row and allocates them
under a common storage budget. It alternates between selecting row candidates
under a fixed probability model and updating that model from the selected
fields. Each row's $K$ candidates use scales $\rho_k s$, with ratios in
the empirically chosen range $[0.70,1.45]$.
Starting from $\rho=1$, we add candidates by
bisecting the widest remaining interval in log scale, so candidate sets are
nested as $K$ increases.

Each candidate is quantized at $\rho_k s$ and refit using
\eqref{eq:scale_fit}. Let $D_{r,k}$ be its sum of squared reconstruction
errors on row $r$. All rows initially use the baseline at scale $s$.
To compare candidates under a shared coding model, we estimate
conditional probabilities from the current row selections, adding
$\tfrac12$ to each symbol count before normalization over the coordinate
ranges spanned by all candidates. Writing $(c_v^{(k)},a_v^{(k)})$ for the
fields of candidate $k$, its estimated code length on row $r$ is
\begin{equation}
\label{eq:rdo_rate}
 R_{r,k}=\sum_{v\in r}\Bigl[-\log_2 q_\theta(c_v^{(k)})-\sum_{j=1}^{8}\log_2 q_\theta(a_{v,j}^{(k)}\mid c_v^{(k)})\Bigr].
\end{equation}

We use the baseline's estimated stored size, $B$ bytes, as the allocation
target. Let $\widehat B_{\mathrm{mix}}$ be the estimated stored bytes for
current row selections $k(r)$. The next code-length budget is
\begin{equation}
\label{eq:rdo_budget}
 R_{\mathrm{budget}}=\sum_r R_{r,k(r)}+8\bigl(B-\widehat B_{\mathrm{mix}}\bigr).
\end{equation}
The correction converts the current storage surplus or deficit into bits:
an oversized mixture reduces the next code-length budget, while spare
storage increases it.

For a rate penalty $\lambda\ge0$, each row selects
\begin{equation}
\label{eq:rdo_choice}
 \kappa_\lambda(r)=\arg\min_k\bigl[D_{r,k}+\lambda R_{r,k}\bigr].
\end{equation}
We search for $\lambda$ to keep the modeled code length within
$R_{\mathrm{budget}}$, then update probabilities from the selected fields.
This allocation--update procedure runs for $T$ sweeps.
Further details are given in Appendix~\ref{app:rdo-algorithm}.

\subsection{RANS encoding and the stored representation}
\label{sec:rans-encode}

After scale selection, refitting, and any RDO refinement, the quantized
fields and reconstruction scales are fixed. We encode these fields with
rANS to form the codestream in Figure~\ref{fig:pipeline}.

\paragraph{Frequency tables and tiles.}
We recount the final full-tensor fields and construct seventeen shared integer
frequency tables using the model in \S\ref{sec:entropy}. Each
coordinate table stores a minimum-value offset for zero-based symbol indices.
For parallel decoding, we partition the fields into independent tiles of
$V$ lattice vectors, each containing $9V$ symbols, and use $J=32$ interleaved
rANS states per tile. This layout assigns one tile to a GPU warp and one
state to each lane.

\paragraph{Encoding the fields.}
For a symbol with integer frequency $f$ and cumulative frequency $F$ in a
table totaling $M=2^b$, rANS updates the coder state $\xi$ as follows \citep{duda_ans}
\begin{equation}
\label{eq:rans_encode}
 \xi'=M\left\lfloor\xi/f\right\rfloor+(\xi\bmod f)+F.
\end{equation}
Before this update, renormalization emits payload words as needed to bound
the state. Since decoding reverses the updates, we encode each vector's
coordinate fields in reverse order, then its coset. The decoder recovers
the coset first to select the conditional coordinate tables.

\paragraph{Stored representation.}
The emitted words form each tile's payload. We concatenate these payloads
and store their offsets and terminal rANS states for independent decoding.
The compressed representation also contains the frequency tables, alphabet
offsets, row reconstruction scales, and tensor
layout. Their combined byte count gives the actual stored rate in
\eqref{eq:objective}.
Larger tiles amortize the states and offsets over more weights but expose
less decoding parallelism.

\subsection{GPU decoding and weight reconstruction}
\label{sec:gpu}
\label{sec:recon}
\label{sec:dtype}

Given the stored representation, the decoder follows the lower path of
Figure~\ref{fig:pipeline}. One GPU warp loads a tile's terminal states and
locates its payload using the stored offsets, with one state per lane.
For a current state $\xi'$, it identifies the symbol whose cumulative-frequency
interval contains $u=\xi'\bmod M$ and inverts \eqref{eq:rans_encode}:
\begin{equation}
\label{eq:rans_decode}
 \xi=f\left\lfloor\xi'/M\right\rfloor+u-F.
\end{equation}
Renormalization consumes payload words as required to restore the state.
Each lane decodes a vector's coset before its coordinate fields, using the
coset to choose the conditional tables. Thus entropy decoding recovers the
field tuple $(c,z_1,\ldots,z_7,m)$.

The lane then applies the algebraic inverse in \eqref{eq:point_inverse} to
recover the lattice point and multiplies it by the stored row scale.
Let $\bar\sigma_r$ denote that scale read in FP32. Rounding and saturation
to the requested dtype $\tau$ produce the reconstructed weight:
\begin{equation}
\label{eq:recon}
 \hat W_{rj}=\operatorname{snap}_{\tau}(\bar\sigma_r p_{rj}).
\end{equation}
These steps fuse symbol recovery, lattice reconstruction, and row rescaling
without materializing an intermediate symbol tensor or consulting a
reconstruction codebook. The field representation and entropy coder are
shared across supported floating-point and integer output dtypes.

\section{Experiments}
\label{sec:experiments}

\subsection{Experimental setup}
\label{sec:setup}

We compress linear weights of the image generation model Z-Image-Turbo
\citep{z_image} and the language model Qwen3.8-27B \citep{qwen3_8_27b}.
Diffusion models run in DiffSynth-Studio \citep{diffsynth_studio}.

Evaluation and timing details are provided in Appendix~\ref{sec:eval-details}.
In Appendix~\ref{sec:exp-minimax}, we extend our evaluation of EntroPack to
Qwen-Image-2.1-Turbo \citep{qwen_image_2_1_turbo} for image generation,
Muse-Glimmer-30B \citep{muse_glimmer_30b} and Qwen3.5-9B \citep{qwen3_5} for language modeling,
and MiniMax-H3 \citep{minimax_fl2va} for joint audio and video generation.
Appendix~\ref{sec:exp-ablations} presents our main component ablation. We vary individual
components of weight representation, rate search, and post-search refinement to measure their
effects on achieved bitrate, reconstruction error, and encoding time. For scale search, we measure the speedup from GPU
size estimation (\ref{sec:encoder-acceleration}), compare predicted and actual rANS
payload lengths (\ref{sec:codelength-estimate}), and vary the number of sampled rows
to quantify the trade-off between encoding time and target-rate accuracy (\ref{sec:subsampled-rate-search}).
We examine how much per-row refinement reduces error at a fixed storage budget (\ref{sec:exp-rdo-net}),
how tile size affects decoding throughput and metadata size (\ref{sec:exp-cost}),
and how much decoding time kernel fusion saves (\ref{sec:gpu-fusion}).
Appendix~\ref{sec:runtime-deployment} measures the cost of weight reconstruction in diffusion and LLM
inference and shows how fitting the MiniMax-H3 denoiser within a 24~GiB weight budget avoids CPU transfers.
Appendix~\ref{sec:applications} explores FP8 and INT8 computation with compressed weights
and layer-wise bitrate allocation.

\paragraph{Metrics.}
We report stored rate, relative $L_2$ weight error, task fidelity, model
conversion time, peak GPU memory, and inference time. Stored rate in bpp is eight times the
total stored bytes, including side information, divided by the total number
of weight elements across the evaluated layers.
Linear biases are excluded from the denominator.
Network weight error follows \eqref{eq:objective},
computed jointly over these weights and reported as a percentage.
We evaluate Z-Image-Turbo on GenEval \citep{geneval} and DPG-Bench
\citep{ella_dpg} to measure how weight compression affects prompt following and visual quality.
GenEval tests whether the requested objects appear in the specified numbers,
colors, and positions, while DPG-Bench checks scene details, attributes, and
relationships specified in longer prompts. We generate four images per prompt
across the two benchmarks, for a total of 6,472 images per configuration.
On these same images, we compute PSNR and SSIM against the corresponding
uncompressed BF16 outputs and use PickScore \citep{pickapic} to estimate
human preference. For the language model,
we report perplexity on WikiText-2 \citep{wikitext2}
and accuracy on MMLU \citep{mmlu}, GSM8K
\citep{gsm8k}, HellaSwag \citep{hellaswag}, and ARC-Challenge \citep{arcchallenge}.
We follow the official scoring guidelines for each benchmark.

\paragraph{Baselines.}
We compare with NF4 and FP4 from bitsandbytes (bnb) \citep{qlora},
INT8, FP8, NVFP4, and MXFP4 from torchao \citep{torchao,fp8_formats,mx_formats,nvfp4_format},
and INT8 with convolutional rotation and FP8 from comfy-kitchen (ck) \citep{comfyui}.
We also evaluate the released HyperQuant, HRTN, and SDNQ implementations
\citep{hyperquant,hrtn,sdnq}. Weights that a format cannot represent remain
at BF16 and contribute to its reported storage.

\subsection{Rate--distortion and target-rate accuracy}
\label{sec:exp-dial}

Figure~\ref{fig:rd} plots the aggregate storage and reconstruction error of
Z-Image-Turbo's compressed linear weights. EntroPack consistently achieves
lower weight error than the tested fixed-width formats at comparable storage
rates. Baseline markers are measured at
4- and 8-bit quantization targets. Across the measured targets, achieved rate increases
with the requested rate, while weight error decreases. The absolute relative
targeting error, $100|\Ractual/\Rtgt-1|$, has a mean of 0.80\% and a maximum
of 2.13\%.
At its 4 bpp target, EntroPack stores 4.02 bpp with 7.18\% weight error,
compared with NF4's 4.50 bpp and 9.41\% error.
Entropy-coded alternatives are closer: HyperQuant and
HRTN achieve 7.31\% and 7.38\% error at 4.15 and 4.06 bpp, respectively.

\subsection{Diffusion model results}
\label{sec:exp-e2e}

\begin{table}[!htb]
\centering
\scriptsize
\setlength{\tabcolsep}{1.7pt}
\setlength{\cmidrulekern}{\tabcolsep}
\renewcommand{\arraystretch}{0.95}
\makebox[\linewidth][c]{\resizebox{0.99\linewidth}{!}{%
\begin{tabular}{@{}*{12}{c}@{}}
\toprule
\multirow{2}{*}{Method} & \multicolumn{2}{c}{Storage (bpp)} &
rel.\ $L_2$ & \multirow{2}{*}{PSNR$\uparrow$} & \multirow{2}{*}{SSIM$\uparrow$} &
\multirow{2}{*}{PickScore$\uparrow$} & \multirow{2}{*}{GenEval$\uparrow$} &
\multirow{2}{*}{DPG-Bench$\uparrow$} &
\multicolumn{2}{c}{Time} & Memory \\
\cmidrule(lr){2-3}\cmidrule(lr){10-11}
& Target$\downarrow$ & Actual$\downarrow$ & (\%)$\downarrow$ & & & & & &
Encode (s)$\downarrow$ & Denoise (ms)$\downarrow$ & (GiB)$\downarrow$ \\
\midrule
BF16 & 16 & 16.00 & 0.00 & 99.00 & 1.00 & 22.74 & 0.7667 & 84.99 & 0.00 & 511.29 & 21.42 \\
\midrule
INT8 (torchao) & 8 & 8.01 & 1.08 & 26.33 & 0.91 & 22.74 & 0.7688 & 85.05 & 0.23 & 560.57 & 12.26 \\
FP8 (torchao) & 8 & 8.01 & 2.65 & 22.24 & 0.85 & \underline{22.75} & \underline{0.7694} & \underline{85.08} & 0.16 & 582.85 & 12.24 \\
INT8 convrot (ck) & 8 & 8.01 & \underline{0.97} & 25.32 & 0.90 & 22.74 & 0.7689 & \textbf{85.14} & \underline{0.06} & \underline{465.90} & 12.25 \\
FP8 (ck) & 8 & 8.00 & 2.67 & 19.65 & 0.79 & 22.75 & 0.7677 & 84.98 & \textbf{0.05} & \textbf{349.47} & 12.24 \\
EntroPack(Ours) & 8 & 8.06 & \textbf{0.54} & \textbf{28.49} & \textbf{0.93} & 22.74 & \textbf{0.7704} & 84.95 & 1.76 & 561.04 & 12.41 \\
EntroPack(Ours) & 7 & 7.07 & 0.99 & \underline{26.39} & \underline{0.91} & 22.74 & 0.7650 & 85.01 & 1.36 & 556.97 & \underline{11.21} \\
EntroPack(Ours) & 6 & 6.06 & 1.89 & 23.85 & 0.87 & \textbf{22.75} & 0.7662 & 84.75 & 1.22 & 553.72 & \textbf{10.05} \\
\midrule
NF4 (bnb) & 4 & 4.50 & 9.41 & 17.07 & 0.71 & 22.67 & 0.7603 & 84.85 & \textbf{0.10} & \textbf{523.88} & 8.23 \\
FP4 (bnb) & 4 & 4.50 & 12.73 & 15.88 & 0.68 & 22.69 & 0.7625 & 85.04 & \underline{0.10} & \underline{525.52} & 8.23 \\
NVFP4 (torchao) & 4 & 4.50 & 9.47 & 17.16 & 0.71 & 22.70 & 0.7648 & \underline{85.12} & 0.56 & 807.85 & 8.23 \\
MXFP4 (torchao) & 4 & 4.25 & 12.02 & 16.19 & 0.68 & 22.63 & 0.7539 & 84.41 & 0.56 & 804.38 & 7.97 \\
HRTN & 3.97 & 4.02 & \underline{7.56} & \underline{18.01} & \underline{0.74} & 22.69 & \textbf{0.7723} & 84.75 & 33.23 & 576.11 & 7.77 \\
HyperQuant & 3.87 & 4.02 & 8.02 & 17.65 & 0.73 & \textbf{22.78} & 0.7686 & \textbf{85.15} & 34.25 & 1049.72 & 7.72 \\
SDNQ & 4 & 4.50 & 11.43 & 16.10 & 0.68 & 22.68 & \underline{0.7689} & 84.81 & 0.24 & 529.64 & 8.23 \\
EntroPack(Ours) & 4 & 4.02 & \textbf{7.18} & \textbf{18.13} & \textbf{0.74} & \underline{22.73} & 0.7671 & 85.11 & 1.12 & 550.92 & \underline{7.69} \\
EntroPack(Ours) & 3 & 3.02 & 14.35 & 15.24 & 0.65 & 22.66 & 0.7548 & 85.08 & 1.12 & 548.38 & \textbf{6.53} \\
\bottomrule
\end{tabular}%
}}
\caption{Weight compression and image quality for the Z-Image-Turbo denoiser, with inference time per step and peak memory for the full pipeline.
For quality comparisons, we tune the HRTN and HyperQuant targets to match the actual storage rate of EntroPack at its 4~bpp setting.
Best values are \textbf{bold} and second-best values are \underline{underlined}.}
\label{tab:e2e}
\end{table}

Table~\ref{tab:e2e} compares quantized versions of Z-Image-Turbo with the BF16 model.
To compare methods at similar storage budgets, we group the 8-bit baselines with
EntroPack targets of 6--8~bpp and the 4-bit baselines with EntroPack targets of 3--4~bpp.
Across all methods, GenEval, DPG-Bench, and PickScore remain close to their BF16 values.
All three scores cluster more tightly around BF16 at the higher bitrates.
For example, against a BF16 PickScore of 22.74, the higher-rate configurations span
22.74 to 22.75 and the lower-rate configurations span 22.63 to 22.78.
By these measures, the higher-rate configurations produce images closer overall to the BF16 outputs.

Differences between weight representations are clearer when we compare the weights
and generated images directly with their BF16 counterparts. Among the higher-rate configurations,
EntroPack at 8.06~bpp gives the lowest relative $L_2$ weight error (0.54\%) and the
highest image fidelity (28.49~dB PSNR and 0.93 SSIM). At 7.07~bpp, EntroPack still
ranks second on both image metrics, reproducing the BF16 images more closely than
any of the tested fixed-width eight-bit formats while using fewer bits per weight.

At lower bitrates, the generated images deviate more from their BF16 counterparts,
as reflected in lower PSNR and SSIM. At matched storage near 4.02~bpp, EntroPack has the lowest
weight error (7.18\%) and the highest PSNR (18.13~dB), ahead of HRTN (18.01~dB)
and HyperQuant (17.65~dB). The larger advantage is in encoding and decoding speed.
EntroPack encodes the denoiser in 1.12~s, compared with over 33~s for either baseline.
The breakdown in Table~\ref{tab:zimage-step-breakdown} shows that reconstructing the
weights per denoising step takes 39.20~ms with EntroPack, 68.07~ms with HRTN,
and 539.61~ms with HyperQuant.

\subsection{Language model results}
\label{sec:exp-llm}

\begin{table}[!htb]
\centering
\scriptsize
\setlength{\tabcolsep}{1.4pt}
\setlength{\cmidrulekern}{\tabcolsep}
\renewcommand{\arraystretch}{0.95}
\makebox[\linewidth][c]{\resizebox{0.99\linewidth}{!}{%
\begin{tabular}{@{}*{13}{c}@{}}
\toprule
\multirow{2}{*}{Method} &
\multicolumn{2}{c}{Storage (bpp)} & rel.\ $L_2$ & PPL $\downarrow$ &
\multicolumn{4}{c}{Accuracy (\%) $\uparrow$} &
Time (s) & \multicolumn{2}{c}{Time (ms)} & Memory \\
\cmidrule(lr){2-3}\cmidrule(lr){5-5}\cmidrule(lr){6-9}\cmidrule(lr){10-10}\cmidrule(lr){11-12}
& Target$\downarrow$ & Actual$\downarrow$ & (\%) $\downarrow$ & WikiText-2 &
MMLU & GSM8K & HellaSwag & ARC-C & Encode &
32k prefill & 8-token decode & (GiB) \\
\midrule
BF16 & 16 & 16.00 & 0.00 & 6.16 & 81.78 & 92.87 & 81.74 & 57.17 & 0.00 & 14872.21 & 115.91 & 52.32 \\
\midrule
INT8 (torchao) & 8 & 8.00 & 1.08 & 6.19 & 81.68 & 92.87 & 81.66 & 56.91 & 0.64 & 15490.48 & 238.06 & 29.75 \\
FP8 (torchao) & 8 & 8.01 & 2.65 & \underline{6.14} & \underline{81.89} & 92.57 & 81.67 & \textbf{57.59} & 0.47 & 15148.74 & 429.89 & 29.75 \\
INT8 convrot (ck) & 8 & 8.01 & 0.99 & 6.16 & 81.35 & 92.19 & 81.72 & \underline{57.59} & \underline{0.21} & \underline{12944.05} & \underline{121.55} & 29.74 \\
FP8 (ck) & 8 & 8.00 & 2.67 & 6.16 & \textbf{81.93} & 93.03 & 81.52 & 57.42 & \textbf{0.10} & \textbf{9479.62} & \textbf{112.77} & 29.73 \\
EntroPack & 8 & 8.08 & \textbf{0.54} & 6.16 & 81.78 & 92.87 & \textbf{81.79} & 57.17 & 4.56 & 15031.01 & 302.22 & 30.06 \\
EntroPack & 7 & 7.10 & \underline{0.98} & 6.15 & 81.75 & \textbf{93.10} & 81.70 & 57.34 & 3.52 & 15016.76 & 283.44 & \underline{27.24} \\
EntroPack & 6 & 6.08 & 1.89 & \textbf{6.13} & 81.57 & \underline{93.03} & \underline{81.72} & 56.31 & 2.67 & 15015.37 & 275.50 & \textbf{24.34} \\
\midrule
NF4 (bnb) & 4 & 4.50 & 9.28 & 6.28 & 79.16 & 93.33 & 81.45 & \textbf{58.28} & \textbf{0.26} & \textbf{14919.06} & \textbf{171.05} & 19.77 \\
FP4 (bnb) & 4 & 4.50 & 12.41 & 6.48 & 79.69 & \textbf{95.15} & 81.36 & 56.74 & \underline{0.33} & \underline{14925.10} & \underline{171.09} & 19.77 \\
NVFP4 (torchao) & 4 & 4.50 & 9.49 & 6.45 & \textbf{82.87} & 92.65 & \textbf{81.65} & \underline{57.68} & 1.92 & 16027.27 & 1418.11 & 21.00 \\
MXFP4 (torchao) & 4 & 4.25 & 11.80 & 6.30 & 79.75 & 92.72 & 80.91 & 56.57 & 1.92 & 15995.37 & 1397.78 & 20.27 \\
HRTN & 3.98 & 4.03 & \underline{7.53} & \textbf{6.19} & 80.67 & 93.56 & \underline{81.49} & 57.42 & 104.95 & 15137.84 & 415.74 & \underline{18.48} \\
HyperQuant & 3.95 & 4.03 & 7.58 & \underline{6.21} & 81.94 & 92.80 & 81.49 & 57.17 & 152.68 & 24015.86 & 2790.90 & 19.98 \\
SDNQ & 4 & 4.50 & 11.06 & 6.50 & 77.32 & 92.95 & 81.27 & 56.83 & 0.85 & 14934.07 & 191.97 & 19.85 \\
EntroPack & 4 & 4.03 & \textbf{7.19} & 6.25 & \underline{82.44} & \underline{93.63} & 81.47 & 56.06 & 2.49 & 14975.83 & 236.04 & 18.51 \\
EntroPack & 3 & 3.03 & 14.37 & 6.35 & 81.56 & 90.83 & 79.93 & 56.23 & 2.46 & 14968.31 & 228.59 & \textbf{15.67} \\
\bottomrule
\end{tabular}%
}}
\caption{Weight compression, language model quality, and runtime on Qwen3.8-27B.
Memory reports peak GPU usage during an 8k-token prefill.
For quality comparisons, we tune the HRTN and HyperQuant targets to match the actual storage rate of EntroPack at its 4~bpp setting.
Best values are \textbf{bold} and second-best values are \underline{underlined}.}
\label{tab:llm}
\end{table}

Table~\ref{tab:llm} compares quantized versions of Qwen3.8-27B with BF16 on
WikiText-2 and the other downstream tasks. As with Z-Image-Turbo, task scores stay closer
to BF16 in the higher-rate group. WikiText-2 perplexity ranges from 6.13 to 6.19 in
this group, compared with 6.16 for BF16, and rises as high as 6.50 at lower rates.

EntroPack achieves the two lowest relative $L_2$ errors in the higher-rate group,
0.54\% at 8.08~bpp and 0.98\% at 7.10~bpp. The 7.10~bpp configuration thus preserves
the weights more accurately than any tested eight-bit baseline while using less storage. At matched storage near
4.03~bpp, EntroPack again gives the lowest error, 7.19\% compared with 7.53\%
for HRTN and 7.58\% for HyperQuant.

As on Z-Image-Turbo, EntroPack encodes the weights faster than HRTN and
HyperQuant at the 4~bpp setting. In a 32k-token prefill, all input tokens share
the cost of weight reconstruction and the
runtime stays close to BF16. The relative overhead is larger in the eight-token
decode workload than in diffusion denoising. For latency-sensitive autoregressive generation, the cost of
reconstructing weights on each forward pass can limit the benefit of compression
when only a few tokens share those weights. Appendix~\ref{sec:runtime-deployment}
gives a detailed breakdown of inference time.

\section{Conclusion}
\label{sec:conclusion}

We presented EntroPack, a calibration-free weight compression method that
combines arbitrary target bitrates with efficient GPU reconstruction.
Lattice quantization and conditional entropy modeling define the coded
representation, and sampled storage estimates guide rate selection without
repeated full-stream encoding. The resulting fields are encoded in
independent rANS tiles, allowing symbol decoding and weight reconstruction
to be fused on the GPU without intermediate symbol tensors or reconstruction
codebooks. By separating storage bitrate from numerical precision, this
representation supports floating-point and integer compute formats.
The same codec can also compress FP8- or INT8-quantized weights and
reconstruct them for low-precision inference. Online decoding time grows
with weight count, making the method well suited to compute-intensive
workloads such as diffusion denoising and Transformer prefill. Experiments
in these settings show accurate bitrate control, fast encoding, and lower
weight error than the tested fixed-width formats at comparable storage,
with modest inference overhead.

\section*{AI Use Statement}

We used generative AI tools to assist with parts of the algorithm
implementation and runtime optimization, literature retrieval, review of
theoretical claims, and manuscript editing. The authors are responsible for the final
implementation, experimental results, and content of the paper.

\bibliography{references}
\bibliographystyle{iclr2027_conference}

\clearpage
\appendix
\raggedbottom
\section*{Appendix}

Appendix~\ref{app:experiments} provides evaluation details, additional
image, language, and audio--video results, pipeline ablations, and analyses of rate selection,
weight reconstruction, and inference cost. Appendix~\ref{sec:applications}
examines compressed storage with FP8 and INT8 computation and the allocation
of storage across layers. Appendix~\ref{app:method-details} specifies the
scale-search settings, RDO algorithm, and tiled rANS implementation.
Appendix~\ref{sec:limitations} discusses limitations and future directions.

\section{Supplementary experiments}
\label{app:experiments}

The supplementary experiments examine how EntroPack's accuracy and
computational cost arise from its representation, rate search, and GPU
implementation. Beyond the additional models, we study individual pipeline
components, the accuracy and cost of rate estimation, and the effects of
row refinement, tile parallelism, and kernel fusion. We then relate weight
reconstruction to model-level inference across workloads and memory
budgets, and test compression across numerical dtypes.

\subsection{Evaluation details}
\label{sec:eval-details}

\paragraph{Models and compressed weights.}
For each model, all methods use the same checkpoint and target weight
population. Storage and weight error cover 276 linear-weight tensors in
Z-Image-Turbo, 232 in Qwen-Image-2.1-Turbo, and 259 in MiniMax-H3.
For language models, we evaluate 400 text-backbone matrices in
Qwen3.8-27B, 416 in Muse-Glimmer-30B, and 200 in Qwen3.5-9B, excluding
token embeddings and output projections. Stored rate includes payloads,
probability tables, scales, and decoding metadata. Both rate and relative
$L_2$ error are aggregated over weight elements as in
\eqref{eq:objective}. EntroPack uses the search settings
in Appendix~\ref{app:search}; RDO is disabled except where explicitly enabled
in the ablations. Single-layer experiments use the $10240\times3840$
Z-Image-Turbo weight \texttt{noise\_refiner.0.feed\_forward.w1}, referred
to as the reference layer.

\paragraph{Image generation.}
We evaluate both image models on the complete official prompt sets of
GenEval \citep{geneval} and DPG-Bench \citep{ella_dpg} in
Tables~\ref{tab:e2e} and~\ref{tab:qwen-image}. We generate images at
$512\times512$ for GenEval and $1024\times1024$ for DPG-Bench.
Both models use eight denoising steps
and guidance scale 1.0. Four seeds per prompt
yield 6,472 images per configuration. PSNR and SSIM compare each image
with the BF16 output from the same initial latent; identical images are
assigned a PSNR of 99~dB. PickScore uses the prompt--image logit before
softmax \citep{pickapic}. For these three metrics, we average seeds within
each prompt and then average across prompts from both benchmarks.
Benchmark scores follow the official evaluation procedures.
The application examples in Appendix~\ref{sec:applications} use eight
denoising steps, with compressed and BF16 outputs paired by prompt and seed.

\paragraph{Video and audio generation.}
We evaluate MiniMax-H3 on VBench \citep{vbench}, using the official
prompt sets and scoring procedures for the six dimensions reported in
Table~\ref{tab:minimax}. Each prompt uses five seeds, generating
$480\times832$ videos with 124 frames, 20 denoising steps, 24 frames per
second, and 32~kHz audio. Video SSIM and audio fidelity reuse the outputs
generated for VBench's subject-consistency evaluation, paired with BF16
outputs under the same prompts, seeds, and generation settings.
Video SSIM is averaged over frames. Audio fidelity is the relative $L_2$
error between 80-band log-mel spectrograms of the matched mono waveforms,
using a 1,024-sample Fourier window and a 256-sample hop. Both fidelity
metrics are averaged over all generated videos.

\paragraph{Language modeling.}
All three language models use the WikiText-2 raw test split
\citep{wikitext2}, tokenized with their respective checkpoint tokenizers
in Transformers \citep{transformers}. We use contexts of at most
2,048 tokens with a stride of 512. Overlapping tokens supply context
but are excluded from the loss, following the Transformers sliding-window
scoring principle.\footnote{\url{https://huggingface.co/docs/transformers/perplexity}}
Each predictable token is scored once, including the final incomplete
stride; perplexity is the exponential of total negative log-likelihood
divided by the number of scored tokens. Comparisons use each model's own
BF16 reference. MMLU uses five-shot prompts from the subject-specific
development sets and reports test-set accuracy
\citep{mmlu}. GSM8K uses eight-shot chain-of-thought prompts, greedy
generation with up to 512 new tokens, and exact match on the extracted
numeric answer on the test set \citep{gsm8k}. HellaSwag
\citep{hellaswag} and ARC-Challenge \citep{arcchallenge} use zero-shot
prompts and token-length-normalized answer likelihoods on their validation
and test sets, respectively.

\paragraph{Timing.}
Runtime and peak-memory measurements use PyTorch \citep{pytorch} on one
NVIDIA H20. Encoding time covers the complete compression pipeline.
Inference times in the model-comparison tables and GPU time breakdowns
are computed by summing kernel durations, including online weight
reconstruction where required.
Both image models are timed at $1024\times1024$ with the prompt shown below.
MiniMax-H3 timing uses 120 frames at $768\times1344$.
In Table~\ref{tab:llm}, batch size is one. Prefill processes 32,768 tokens.
The eight-token workload uses a multi-token
prediction (MTP)-style draft-and-verify cycle after a 32k-token prefix,
reporting time per eight tokens under full acceptance.
The workload sweeps in Appendix~\ref{sec:runtime-deployment} vary
batch size, positions per forward pass, and prefill length separately.

\smallskip
\begingroup
\setlength{\fboxsep}{6pt}
\noindent\fbox{\begin{minipage}{\dimexpr\linewidth-2\fboxsep-2\fboxrule\relax}
\small
\textbf{Image timing prompt.}\par\smallskip
Young Chinese woman in red Hanfu, intricate embroidery. Impeccable
makeup, red floral forehead pattern. Elaborate high bun, golden phoenix
headdress, red flowers, beads. Holds round folding fan with lady, trees,
bird. Neon lightning-bolt lamp (\Lightning), bright yellow glow, above
extended left palm. Soft-lit outdoor night background, silhouetted tiered
pagoda (\begin{CJK}{UTF8}{gbsn}西安大雁塔\end{CJK}), blurred colorful distant lights.
\end{minipage}}\par
\endgroup
\smallskip

\paragraph{Peak memory and weight residency.}
Tables~\ref{tab:e2e} and~\ref{tab:qwen-image}
report peak allocated GPU memory over the full image request, including
text encoding, denoising, and VAE decoding, with eligible linear weights
compressed throughout the pipeline. Table~\ref{tab:minimax} reports the
MiniMax-H3 DiT-forward peak, and Table~\ref{tab:llm} reports the peak during
an 8k-token prefill at batch size one. Peak memory includes model weights
and runtime allocations.

\FloatBarrier
\subsection{Additional Results}
\label{sec:exp-minimax}
\label{sec:additional-results}

\paragraph{Image generation with Qwen-Image-2.1-Turbo.}
Table~\ref{tab:qwen-image} extends the image evaluation to
Qwen-Image-2.1-Turbo \citep{qwen_image_2_1_turbo} using the same benchmarks
as Table~\ref{tab:e2e}. The results follow the pattern observed for
Z-Image-Turbo in Section~\ref{sec:exp-e2e}: at higher bitrates, GenEval,
DPG-Bench, and PickScore remain close to BF16, while weight error, PSNR,
and SSIM distinguish the representations more clearly. EntroPack at the
8~bpp target gives the lowest weight error and the highest image fidelity.
Its 7~bpp configuration again ranks second on PSNR and SSIM while using
less storage than the eight-bit baselines.

At the 4~bpp target, EntroPack also has the lowest weight error, with
PSNR and SSIM comparable to HRTN and HyperQuant at matched storage.
As on Z-Image-Turbo, the clearer advantage over these entropy-coded
baselines is in encoding cost: EntroPack takes 1.16~s, compared with
over 33~s for either baseline. It also substantially reduces peak memory
relative to BF16 while keeping denoising time close to the uncompressed
model.

\begin{table}[!htb]
\centering
\scriptsize
\setlength{\tabcolsep}{1.7pt}
\setlength{\cmidrulekern}{\tabcolsep}
\renewcommand{\arraystretch}{0.98}
\makebox[\linewidth][c]{\resizebox{0.99\linewidth}{!}{%
\begin{tabular}{@{}*{12}{c}@{}}
\toprule
\multirow{2}{*}{Method} & \multicolumn{2}{c}{Storage (bpp)} &
rel.\ $L_2$ & \multirow{2}{*}{PSNR$\uparrow$} & \multirow{2}{*}{SSIM$\uparrow$} &
\multirow{2}{*}{PickScore$\uparrow$} & \multirow{2}{*}{GenEval$\uparrow$} &
\multirow{2}{*}{DPG-Bench$\uparrow$} & \multicolumn{2}{c}{Time} & Memory \\
\cmidrule(lr){2-3}\cmidrule(lr){10-11}
& Target$\downarrow$ & Actual$\downarrow$ & (\%)$\downarrow$ & & & & & &
Encode (s)$\downarrow$ & Denoise (ms)$\downarrow$ & (GiB)$\downarrow$ \\
\midrule
BF16 & 16 & 16.00 & 0.00 & 99.00 & 1.00 & 22.31 & 0.4931 & 83.32 & 0.00 & 554.01 & 33.98 \\
\midrule
INT8 (torchao) & 8 & 8.01 & 1.10 & 29.24 & 0.94 & 22.31 & 0.4939 & 83.23 & 0.22 & 609.42 & 19.79 \\
FP8 (torchao) & 8 & 8.01 & 2.65 & 24.80 & 0.90 & 22.29 & 0.4951 & \underline{83.39} & 0.16 & 636.46 & 21.92 \\
INT8 convrot (ck) & 8 & 8.01 & \underline{0.98} & 27.47 & 0.93 & \underline{22.32} & 0.4941 & 83.37 & \underline{0.06} & \underline{486.55} & 19.79 \\
FP8 (ck) & 8 & 8.00 & 2.68 & 22.41 & 0.86 & \textbf{22.33} & 0.4943 & \textbf{83.61} & \textbf{0.04} & \textbf{356.87} & 19.78 \\
EntroPack & 8 & 8.07 & \textbf{0.54} & \textbf{32.09} & \textbf{0.96} & 22.31 & \textbf{0.4979} & 83.25 & 1.88 & 606.48 & 20.05 \\
EntroPack & 7 & 7.09 & 0.99 & \underline{29.91} & \underline{0.95} & 22.31 & \underline{0.4951} & 83.32 & 1.47 & 601.87 & \underline{18.26} \\
EntroPack & 6 & 6.07 & 1.90 & 26.39 & 0.92 & 22.31 & 0.4938 & 83.20 & 1.27 & 599.26 & \textbf{16.46} \\
\midrule
NF4 (bnb) & 4 & 4.50 & 9.42 & 18.26 & 0.77 & 22.23 & 0.4803 & \textbf{83.55} & \textbf{0.09} & \textbf{567.96} & 13.59 \\
FP4 (bnb) & 4 & 4.50 & 12.77 & 17.01 & 0.73 & 22.22 & 0.4815 & 83.15 & \underline{0.10} & \underline{570.18} & 13.59 \\
NVFP4 (torchao) & 4 & 4.50 & 9.45 & 18.77 & 0.78 & 22.26 & \textbf{0.4890} & 83.06 & 0.58 & 894.94 & 13.59 \\
MXFP4 (torchao) & 4 & 4.25 & 11.97 & 15.94 & 0.70 & 22.22 & 0.4622 & \underline{83.55} & 0.59 & 889.36 & 13.20 \\
HRTN & 3.96 & 4.03 & \underline{7.54} & \underline{19.77} & \textbf{0.81} & \textbf{22.31} & 0.4815 & 83.39 & 33.38 & 628.84 & 13.67 \\
HyperQuant & 3.94 & 4.03 & 7.62 & \textbf{19.78} & \underline{0.81} & \underline{22.28} & 0.4816 & 83.15 & 38.82 & 1159.68 & 13.97 \\
SDNQ & 4 & 4.50 & 11.53 & 18.02 & 0.76 & 22.20 & 0.4680 & 83.01 & 0.26 & 572.72 & 13.59 \\
EntroPack & 4 & 4.03 & \textbf{7.19} & 19.75 & 0.81 & 22.28 & \underline{0.4843} & 83.35 & 1.16 & 595.41 & \underline{12.74} \\
EntroPack & 3 & 3.02 & 14.35 & 16.19 & 0.70 & 22.11 & 0.4474 & 82.65 & 1.14 & 591.68 & \textbf{10.98} \\
\bottomrule
\end{tabular}%
}}
\caption{Weight compression and image quality for the Qwen-Image-2.1-Turbo denoiser, with inference time per step and peak memory for the full pipeline. For quality comparisons, HRTN and HyperQuant targets are adjusted to match EntroPack at its 4~bpp setting. Best values are \textbf{bold} and second-best values are \underline{underlined}.}
\label{tab:qwen-image}
\end{table}

\FloatBarrier
\paragraph{Language modeling with Muse-Glimmer-30B and Qwen3.5-9B.}
Table~\ref{tab:additional-llms} evaluates Muse-Glimmer-30B
\citep{muse_glimmer_30b} and Qwen3.5-9B \citep{qwen3_5} on the five
language benchmarks used in Table~\ref{tab:llm}. Across both models,
higher-rate configurations largely retain BF16 task performance,
consistent with Section~\ref{sec:exp-llm}. EntroPack gives the lowest
weight error in both bitrate groups at its 8 and 4~bpp targets. Its
advantage over HRTN and HyperQuant at matched storage therefore extends
to both additional models, although the downstream task rankings differ.

The effect of lower bitrates on task accuracy depends on the model.
Muse-Glimmer-30B remains close to BF16 across the five benchmarks at the
4~bpp target, whereas Qwen3.5-9B is more sensitive to further compression.
Reducing EntroPack's target from 4 to 3~bpp lowers Qwen3.5-9B's GSM8K
accuracy from 85.90\% to 69.67\%, with declines on MMLU and HellaSwag
as well. Perplexity decreases at this setting despite the loss in task
accuracy, illustrating why the downstream benchmarks are needed
alongside weight error and perplexity.

\begin{table}[!htb]
\centering
\scriptsize
\setlength{\tabcolsep}{1.4pt}
\setlength{\cmidrulekern}{\tabcolsep}
\renewcommand{\arraystretch}{0.95}
\makebox[\linewidth][c]{\resizebox{0.99\linewidth}{!}{%
\begin{tabular}{@{}*{13}{c}@{}}
\toprule
\multirow{2}{*}{Method} & \multicolumn{2}{c}{Storage (bpp)} & rel.\ $L_2$ & PPL$\downarrow$ & \multicolumn{4}{c}{Accuracy (\%)$\uparrow$} &
Time (s) & \multicolumn{2}{c}{Time (ms)} & Memory \\
\cmidrule(lr){2-3}\cmidrule(lr){5-5}\cmidrule(lr){6-9}\cmidrule(lr){10-10}\cmidrule(lr){11-12}
& Target$\downarrow$ & Actual$\downarrow$ & (\%)$\downarrow$ & WikiText-2 & MMLU & GSM8K & HellaSwag & ARC-C &
Encode & 32k prefill & 8-token decode & (GiB) \\
\midrule
\multicolumn{13}{c}{\textbf{Muse-Glimmer-30B}} \\
\midrule
BF16 & 16 & 16.00 & 0.00 & 5.63 & 81.29 & 93.03 & 80.63 & 63.14 & 0.00 & 19063.30 & -- & 53.80 \\
\midrule
INT8 (torchao) & 8 & 8.00 & 1.05 & \underline{5.63} & 81.13 & \underline{93.48} & 80.61 & 63.14 & 0.68 & 19590.07 & -- & 30.34 \\
FP8 (torchao) & 8 & 8.00 & 2.65 & 5.63 & 81.19 & 93.18 & \textbf{80.77} & 63.14 & 0.51 & 19353.06 & -- & 30.96 \\
INT8 convrot (ck) & 8 & 8.00 & 1.00 & 5.63 & 81.12 & 92.80 & 80.55 & \underline{63.40} & \underline{0.21} & \underline{17066.79} & -- & 30.33 \\
FP8 (ck) & 8 & 8.00 & 2.67 & 5.64 & 80.84 & 93.33 & 80.52 & 62.63 & \textbf{0.13} & \textbf{13551.62} & -- & 30.32 \\
EntroPack(Ours) & 8 & 8.08 & \textbf{0.54} & \textbf{5.63} & \textbf{81.28} & 92.87 & 80.55 & \textbf{63.48} & 4.58 & 19237.40 & -- & 30.63 \\
EntroPack(Ours) & 7 & 7.10 & \underline{0.98} & 5.63 & \underline{81.28} & \textbf{93.48} & \underline{80.71} & 63.31 & 3.40 & 19175.16 & -- & \underline{27.70} \\
EntroPack(Ours) & 6 & 6.08 & 1.88 & 5.64 & 81.19 & 92.95 & 80.56 & 63.05 & 2.84 & 19068.27 & -- & \textbf{24.71} \\
\midrule
NF4 (bnb) & 4 & 4.50 & 9.25 & 5.76 & 80.36 & 92.19 & 80.20 & 61.26 & \textbf{0.35} & \textbf{19100.32} & -- & 20.08 \\
FP4 (bnb) & 4 & 4.50 & 12.34 & 5.91 & 79.58 & 90.83 & 79.16 & 61.69 & \underline{0.41} & 19105.69 & -- & 20.08 \\
NVFP4 (torchao) & 4 & 4.50 & 9.50 & 5.77 & 80.27 & 92.34 & 80.03 & 62.71 & 2.05 & 20256.04 & -- & 23.18 \\
MXFP4 (torchao) & 4 & 4.25 & 11.76 & 5.90 & 79.57 & 89.46 & 79.65 & 61.43 & 2.05 & 20163.88 & -- & 22.43 \\
HRTN & 3.98 & 4.03 & \underline{7.54} & \textbf{5.73} & 80.37 & \underline{92.42} & 80.23 & 61.18 & 108.60 & 19183.84 & -- & 35.97 \\
HyperQuant & 3.95 & 4.03 & 7.59 & 5.75 & \textbf{80.89} & \textbf{92.95} & \textbf{80.34} & \textbf{63.14} & 83.20 & 29281.86 & -- & 20.76 \\
SDNQ & 4 & 4.50 & 10.96 & 5.89 & 80.32 & 90.60 & 79.85 & 61.18 & 0.90 & 19127.90 & -- & 20.31 \\
EntroPack(Ours) & 4 & 4.03 & \textbf{7.18} & \underline{5.73} & \underline{80.79} & 91.81 & \underline{80.28} & \underline{62.80} & 2.56 & 19109.69 & -- & \underline{18.70} \\
EntroPack(Ours) & 3 & 3.03 & 14.35 & 6.05 & 77.99 & 89.54 & 77.90 & 60.84 & 2.53 & \underline{19102.35} & -- & \textbf{15.98} \\
\midrule
\multicolumn{13}{c}{\textbf{Qwen3.5-9B}} \\
\midrule
BF16 & 16 & 16.00 & 0.00 & 8.77 & 79.84 & 87.57 & 77.50 & 53.75 & 0.00 & 4493.99 & 53.78 & 18.20 \\
\midrule
INT8 (torchao) & 8 & 8.00 & 1.07 & 8.88 & 79.91 & 87.95 & 77.52 & 53.16 & 0.21 & 4686.08 & 92.42 & 11.77 \\
FP8 (torchao) & 8 & 8.01 & 2.65 & 8.56 & 79.84 & 86.28 & \textbf{77.59} & 53.50 & 0.16 & 4546.45 & 153.33 & 11.77 \\
INT8 convrot (ck) & 8 & 8.01 & \underline{0.97} & 8.83 & 79.80 & \textbf{88.40} & \underline{77.57} & 53.24 & \underline{0.08} & \underline{3958.83} & \underline{55.83} & 11.77 \\
FP8 (ck) & 8 & 8.00 & 2.68 & \textbf{8.43} & \textbf{80.03} & 86.96 & 77.56 & \textbf{54.01} & \textbf{0.05} & \textbf{2980.71} & \textbf{53.55} & 11.77 \\
EntroPack(Ours) & 8 & 8.06 & \textbf{0.54} & 8.81 & 79.94 & 87.72 & 77.51 & \underline{53.84} & 1.81 & 4516.21 & 119.82 & 11.87 \\
EntroPack(Ours) & 7 & 7.09 & 0.99 & 8.74 & \underline{79.98} & 87.95 & 77.51 & 53.75 & 1.46 & 4513.37 & 109.57 & \underline{11.06} \\
EntroPack(Ours) & 6 & 6.07 & 1.90 & \underline{8.49} & 79.77 & \underline{88.02} & 77.36 & 52.90 & 1.18 & 4512.28 & 107.78 & \textbf{10.25} \\
\midrule
NF4 (bnb) & 4 & 4.50 & 9.31 & 8.84 & 77.28 & 82.34 & \underline{76.60} & 52.73 & \textbf{0.11} & \textbf{4479.40} & \textbf{65.75} & 8.96 \\
FP4 (bnb) & 4 & 4.50 & 12.48 & 8.88 & 76.19 & 78.39 & 76.45 & \textbf{55.46} & \underline{0.13} & \underline{4481.30} & \underline{65.76} & 8.96 \\
NVFP4 (torchao) & 4 & 4.50 & 9.48 & 9.27 & \underline{78.66} & 83.78 & 76.46 & 53.58 & 0.58 & 4798.44 & 467.87 & 9.74 \\
MXFP4 (torchao) & 4 & 4.25 & 11.84 & 8.94 & 77.16 & 80.82 & 76.39 & 51.96 & 0.58 & 4788.69 & 462.08 & 9.55 \\
HRTN & 3.98 & 4.03 & \underline{7.54} & 9.11 & 78.56 & 85.37 & 76.60 & 53.41 & 34.01 & 4542.97 & 148.92 & 14.41 \\
HyperQuant & 3.95 & 4.03 & 7.60 & \underline{8.24} & \textbf{78.95} & \underline{85.82} & 76.60 & 53.75 & 41.05 & 7843.29 & 873.08 & 9.56 \\
SDNQ & 4 & 4.50 & 11.16 & 8.66 & 77.36 & 78.47 & \textbf{76.92} & 52.05 & 0.26 & 4512.69 & 77.12 & 9.05 \\
EntroPack(Ours) & 4 & 4.03 & \textbf{7.21} & 9.88 & 78.55 & \textbf{85.90} & 76.28 & 53.67 & 1.05 & 4533.06 & 102.57 & \underline{8.58} \\
EntroPack(Ours) & 3 & 3.03 & 14.39 & \textbf{7.87} & 73.64 & 69.67 & 74.15 & \underline{53.92} & 1.06 & 4529.49 & 98.17 & \textbf{7.87} \\
\bottomrule
\end{tabular}%
}}
\caption{Weight compression and language model quality on Muse-Glimmer-30B and Qwen3.5-9B. Dashes indicate that Muse-Glimmer-30B does not support multi-token prediction. Best values are \textbf{bold} and second-best values are \underline{underlined}.}
\label{tab:additional-llms}
\end{table}

\FloatBarrier
\paragraph{Joint video and audio generation with MiniMax-H3.}
Table~\ref{tab:minimax} extends the evaluation to MiniMax-H3
\citep{minimax_fl2va} for joint video and audio generation. As in the
image experiments, VBench \citep{vbench} scores can remain near BF16
even when the generated outputs change, while paired video SSIM and
audio error distinguish their fidelity to the reference. EntroPack at
the 8~bpp target gives the lowest weight and audio errors and the highest
video SSIM among the higher-rate configurations. At the 4~bpp target,
it retains the lowest weight error and highest video SSIM in its group,
with audio error comparable to HRTN and HyperQuant.

The encoding and memory benefits are consistent with the image results,
while the longer video denoising computation amortizes weight
reconstruction more effectively. At the 4~bpp target, EntroPack reduces
peak DiT memory from 67.15~GiB to 21.29~GiB, with denoising time remaining
close to BF16 at 26.34~s versus 26.21~s per step. As on Z-Image-Turbo,
encoding is substantially faster than with HRTN and HyperQuant.

\begin{table}[!htb]
\centering
\scriptsize
\setlength{\tabcolsep}{3pt}
\setlength{\cmidrulekern}{\tabcolsep}
\renewcommand{\arraystretch}{0.98}
\begin{tabular*}{0.99\linewidth}{@{\extracolsep{\fill}}*{9}{c}@{}}
\toprule
\multirow{2}{*}{Method} & \multicolumn{2}{c}{Storage (bpp)} & Weights & Video & Audio & \multicolumn{2}{c}{Time (s)} & DiT memory \\
\cmidrule(lr){2-3}\cmidrule(lr){7-8}
& Target$\downarrow$ & Actual$\downarrow$ & rel.\ $L_2$ (\%)$\downarrow$ & SSIM$\uparrow$ & rel.\ $L_2$ (\%)$\downarrow$ & Encode$\downarrow$ & Denoise$\downarrow$ & (GiB)$\downarrow$ \\
\midrule
BF16 & 16 & 16.00 & 0.00 & 1.00 & 0.00 & 0.00 & 26.21 & 67.15 \\
\midrule
INT8 (torchao) & 8 & 8.00 & 1.04 & 0.71 & 13.14 & 0.87 & 26.86 & 36.58 \\
FP8 (torchao) & 8 & 8.01 & 2.64 & 0.65 & 17.96 & 0.63 & 26.60 & 36.57 \\
INT8 convrot (ck) & 8 & 8.01 & \underline{0.89} & 0.68 & 28.13 & \underline{0.25} & \underline{24.29} & 36.58 \\
FP8 (ck) & 8 & 8.00 & 2.67 & 0.63 & 20.00 & \textbf{0.14} & \textbf{21.23} & 36.54 \\
EntroPack & 8 & 8.09 & \textbf{0.53} & \textbf{0.76} & \textbf{10.60} & 4.70 & 26.43 & 36.96 \\
EntroPack & 7 & 7.11 & 0.96 & \underline{0.72} & \underline{13.06} & 3.85 & 26.42 & \underline{33.16} \\
EntroPack & 6 & 6.08 & 1.84 & 0.68 & 15.53 & 2.79 & 26.38 & \textbf{29.19} \\
\midrule
NF4 (bnb) & 4 & 4.50 & 9.14 & 0.53 & 39.98 & \textbf{0.38} & \textbf{26.27} & 23.09 \\
FP4 (bnb) & 4 & 4.50 & 12.03 & 0.54 & 44.77 & \underline{0.46} & \underline{26.27} & 23.09 \\
NVFP4 (torchao) & 4 & 4.50 & 9.77 & 0.54 & 34.71 & 2.57 & 27.79 & 23.87 \\
MXFP4 (torchao) & 4 & 4.25 & 11.51 & 0.55 & 39.87 & 2.57 & 27.74 & 22.91 \\
HRTN & 3.86 & 4.03 & \underline{7.42} & \underline{0.56} & \underline{30.31} & 136.38 & 26.58 & \underline{21.26} \\
HyperQuant & 3.82 & 4.03 & 8.34 & 0.55 & \textbf{30.24} & 207.15 & 32.31 & 28.69 \\
SDNQ & 4 & 4.50 & 10.24 & 0.50 & 39.24 & 1.17 & 26.35 & 23.08 \\
EntroPack & 4 & 4.03 & \textbf{7.02} & \textbf{0.58} & 30.52 & 2.59 & 26.34 & 21.29 \\
EntroPack & 3 & 3.03 & 14.09 & 0.52 & 52.31 & 2.58 & 26.33 & \textbf{17.41} \\
\bottomrule
\addlinespace[6pt]
\toprule
\multirow{2}{*}{Method} & \multicolumn{2}{c}{Storage (bpp)} & \multicolumn{6}{c}{VBench (\%)$\uparrow$} \\
\cmidrule(lr){2-3}\cmidrule(lr){4-9}
& Target$\downarrow$ & Actual$\downarrow$ & \shortstack{Subject\\consistency} & \shortstack{Motion\\smoothness} & \shortstack{Spatial\\relationship} & \shortstack{Overall\\consistency} & \shortstack{Aesthetic\\quality} & \shortstack{Imaging\\quality} \\
\midrule
BF16 & 16 & 16.00 & 91.45 & 97.59 & 71.34 & 71.93 & 60.77 & 66.17 \\
\midrule
INT8 (torchao) & 8 & 8.00 & 91.28 & 97.57 & \underline{71.70} & 71.60 & 60.84 & 66.05 \\
FP8 (torchao) & 8 & 8.01 & 91.04 & 97.56 & 69.94 & 71.54 & \textbf{60.93} & 65.83 \\
INT8 convrot (ck) & 8 & 8.01 & 91.53 & 97.48 & \textbf{72.86} & \underline{71.64} & 60.85 & \textbf{66.68} \\
FP8 (ck) & 8 & 8.00 & 91.21 & 97.46 & 71.52 & \textbf{71.86} & 60.85 & 66.11 \\
EntroPack & 8 & 8.09 & \textbf{91.56} & \underline{97.61} & 70.84 & 71.53 & \underline{60.87} & 65.93 \\
EntroPack & 7 & 7.11 & \underline{91.54} & 97.56 & 69.61 & 71.60 & 60.84 & \underline{66.19} \\
EntroPack & 6 & 6.08 & 91.38 & \textbf{97.62} & 71.12 & 71.40 & 60.84 & 65.80 \\
\midrule
NF4 (bnb) & 4 & 4.50 & 92.24 & 97.46 & \underline{72.80} & \textbf{71.73} & 60.58 & 66.79 \\
FP4 (bnb) & 4 & 4.50 & 91.09 & 97.39 & 71.16 & 69.33 & 59.41 & 64.78 \\
NVFP4 (torchao) & 4 & 4.50 & 91.13 & 97.28 & 70.84 & 69.96 & 60.64 & 66.46 \\
MXFP4 (torchao) & 4 & 4.25 & \textbf{93.37} & \textbf{98.16} & 67.61 & 69.44 & 59.63 & 61.75 \\
HRTN & 3.86 & 4.03 & 92.13 & \underline{97.74} & 70.48 & 71.23 & 60.77 & 67.25 \\
HyperQuant & 3.82 & 4.03 & 91.44 & 97.54 & \textbf{73.25} & 70.66 & 60.24 & \underline{67.31} \\
SDNQ & 4 & 4.50 & \underline{93.29} & 97.68 & 72.14 & 71.01 & \textbf{61.99} & \textbf{68.09} \\
EntroPack & 4 & 4.03 & 92.14 & 97.68 & 70.98 & \underline{71.58} & \underline{61.11} & 66.24 \\
EntroPack & 3 & 3.03 & 93.09 & 96.82 & 67.24 & 69.09 & 59.37 & 63.52 \\
\bottomrule
\end{tabular*}
\caption{Weight compression, video and audio quality, and denoising time and peak memory for MiniMax-H3. Audio error compares log-mel spectrograms. HRTN and HyperQuant use their nominal 4~bpp settings for video SSIM, audio error, and imaging quality. Best values are \textbf{bold} and second-best values are \underline{underlined}.}
\label{tab:minimax}
\end{table}

\FloatBarrier
\subsection{Ablation of Pipeline Components}
\label{sec:exp-ablations}

EntroPack integrates multiple components for weight representation, rate
search, and post-search refinement. The representation combines $E_8$
lattice quantization, parity reduction, and coset-conditioned entropy
coding; rate search combines row subsampling with analytic size estimation;
and post-search refinement includes row-scale refitting and optional RDO.
We assess the contribution of each design through controlled ablations in
Table~\ref{tab:coding} over all 276
linear-weight tensors of Z-Image-Turbo at target rates of 4, 6, and 8 bpp.

\begin{table}[!htb]
\centering
\footnotesize
\setlength{\tabcolsep}{2.5pt}
\renewcommand{\arraystretch}{1.08}
\resizebox{\linewidth}{!}{%
\begin{tabular}{cccccccccc}
\toprule
\multirow{2}{*}{Variant} & \multicolumn{3}{c}{Target 4 bpp} & \multicolumn{3}{c}{Target 6 bpp} & \multicolumn{3}{c}{Target 8 bpp} \\
\cmidrule(lr){2-4}\cmidrule(lr){5-7}\cmidrule(lr){8-10}
 & Actual bpp & Rel.\ $L_2$ (\%) & Encode (s) & Actual bpp & Rel.\ $L_2$ (\%) & Encode (s) & Actual bpp & Rel.\ $L_2$ (\%) & Encode (s) \\
\midrule
Full EntroPack (RDO off) & 4.0237 & 7.1834 & 1.12 & 6.0555 & 1.8936 & 1.22 & 8.0604 & 0.5415 & 1.76 \\
\midrule
No $E_8$ (scalar quantization) & 4.0271 & 7.7411 & 1.22 & 6.0704 & 2.0427 & 1.30 & 8.1011 & 0.5824 & 1.55 \\
No coset conditioning (pooled tables) & 4.0271 & 7.1996 & 1.12 & 6.0697 & 1.8921 & 1.26 & 8.1045 & 0.5342 & 1.82 \\
No parity reduction (direct $z_8$) & 4.0235 & 7.8233 & 1.34 & 6.0553 & 2.0617 & 1.46 & 8.0605 & 0.5869 & 1.96 \\
\midrule
No subsampling (full-matrix search) & 4.0178 & 7.2122 & 2.74 & 6.0352 & 1.9180 & 3.15 & 8.0691 & 0.5386 & 5.74 \\
No size estimate (full-matrix rANS coding) & 4.0000 & 7.3006 & 6.18 & 5.9999 & 1.9631 & 7.81 & 8.0001 & 0.5620 & 11.45 \\
\midrule
No row-scale refit & 4.0237 & 7.2029 & 1.13 & 6.0555 & 1.8942 & 1.24 & 8.0604 & 0.5416 & 1.78 \\
With RDO & 4.0237 & 7.0668 & 8.14 & 6.0555 & 1.8683 & 8.49 & 8.0604 & 0.5356 & 11.45 \\
\bottomrule
\end{tabular}
}
\caption{Effects of pipeline component ablations on stored rate, relative
$L_2$ reconstruction error, and encoding time for Z-Image-Turbo.}
\label{tab:coding}
\end{table}

\paragraph{Representation and coding.}
We use $E_8$ lattice quantization to represent blocks of eight weights
jointly and reduce reconstruction error within a storage budget. At the
4 bpp target, replacing it with scalar quantization increases error from
7.18\% to 7.74\% despite slightly higher storage. Coset conditioning models
the field distributions separately for the two cosets to capture their
dependence. Pooling these distributions at the same target slightly
increases both stored rate and error. Parity reduction uses the parity
implied by the first seven coordinates to encode the eighth coordinate
more compactly. Removing it means directly encoding the eighth integer
coordinate $z_8$ while retaining the $E_8$ projection. At the 4 bpp target,
the extra bits needed for this coordinate force a coarser quantization
scale, raising error to 7.82\% at a nearly unchanged actual rate. These
changes have little effect on encoding time.

\paragraph{Rate-selection cost.}
At fixed quantization scales, subsampling and size estimation leave the
actual bpp and reconstruction error unchanged. They accelerate scale
selection by approximating candidate rates, at the cost of less precise
matching between target and actual bpp.
We subsample complete rows to limit the data processed at each candidate
scale and avoid repeated full-matrix scans during rate search. At the
4 bpp target, using all rows instead increases model encoding time from
1.12 s to 2.74 s, with only small changes in stored rate and reconstruction
error. We also use analytic size estimation to evaluate candidate rates
without constructing codestreams; actual rANS encoding is performed after
scale selection. With full-matrix search at the 4 bpp target, replacing
this estimate with actual rANS coding increases model encoding time from
2.74 s to 6.18 s. The actual stored rate is 4.0000 bpp, compared with
4.0178 bpp for full-matrix search using the estimate.

\paragraph{Post-search refinement.}
After rate selection, EntroPack refits each row's reconstruction scale
to reduce error while keeping the lattice points and entropy-coded payload
fixed. At the 4 bpp target, omitting this step increases error from 7.18\%
to 7.20\%, with unchanged storage and nearly unchanged encoding time.
RDO provides an optional refinement when additional conversion time is
acceptable. It selects among candidate quantizations, each with its fitted
row scale, to balance distortion and estimated storage across rows. At
4 bpp, enabling RDO reduces error from 7.18\% to 7.07\% at essentially
unchanged actual storage, while increasing encoding time from 1.12~s to
8.14~s.

\FloatBarrier
\subsection{Accelerating Stored-size Estimation}
\label{sec:encoder-acceleration}

\paragraph{Estimation overhead.}
The stored-size estimator evaluates candidate rates from symbol histograms
without constructing codestreams. These statistics are generated on the
GPU, but evaluating them on the CPU requires transferring the arrays for
frequency normalization and code-length computation. Scale
search repeats this process for every candidate and every layer, so the
transfers, synchronization, and small CPU operations accumulate even when
only a subset of rows is sampled.

\paragraph{GPU evaluation.}
To reduce this overhead, EntroPack keeps candidate statistics on the GPU
and performs frequency normalization and code-length evaluation there.
The CPU updates the search interval from the returned stored-size estimate,
without processing histogram arrays at every iteration. We also vectorize
the frequency-normalization work that remains on the CPU for final
probability-table construction. These two changes reduce data movement
and CPU processing while retaining the same estimation formulas,
quantization procedure, and compressed format.

\begin{table}[!htb]
\centering
\footnotesize
\setlength{\tabcolsep}{6pt}
\renewcommand{\arraystretch}{1.08}
\begin{tabular}{ccccc}
\toprule
Model & Target bpp & CPU evaluator (s) & GPU evaluator (s) & Speedup \\
\midrule
\multirow{3}{*}{Z-Image-Turbo} & 4 & 3.29 & 1.12 & $2.95\times$ \\
 & 6 & 6.12 & 1.22 & $5.01\times$ \\
 & 8 & 10.66 & 1.76 & $6.05\times$ \\
\midrule
\multirow{3}{*}{MiniMax-H3} & 4 & 4.87 & 2.59 & $1.88\times$ \\
 & 6 & 7.38 & 2.79 & $2.65\times$ \\
 & 8 & 14.22 & 4.70 & $3.03\times$ \\
\bottomrule
\end{tabular}
\caption{Model encoding time with stored-size estimation on the CPU or GPU.}
\label{tab:encoder-acceleration}
\end{table}

\paragraph{Encoding performance.}
Table~\ref{tab:encoder-acceleration} compares the two implementations with
GPU-resident weights at fixed target rates. At 4 bpp, the GPU
configuration reduces model encoding time from 3.29~s to 1.12~s for
Z-Image-Turbo and from 4.87~s to 2.59~s for MiniMax-H3. All tested
configurations produce byte-identical compressed outputs, with unchanged
actual bpp and reconstructed weights.

\FloatBarrier
\subsection{Analysis of Payload Length Estimation}
\label{sec:codelength-estimate}

The stored-size estimator uses symbol counts to evaluate candidate
quantizations without running the entropy coder. We analyze its payload
term in \eqref{eq:estimate}, whose state correction accounts for
information that remains in the terminal rANS states instead of being
emitted as payload words.

\paragraph{Payload estimation with state correction.}
Encoding symbol $i$ with model probability $p_i$ emits $\ell_i$ bits
and changes its lane state from $\xi_i$ to $\xi_i'$. We define the
accumulated discrepancy of the integer rANS updates, in bits, as
\[
 \varepsilon_{\mathrm{rANS}} :=
 \sum_i \left[\ell_i+\log_2(\xi_i'/\xi_i)-\log_2(1/p_i)\right].
\]
The emitted bits sum to $L_{\mathrm{payload}}$ and the modeled symbol
costs to $C_{\mathrm{CE}}$. Within each lane the log-state changes sum
to $\log_2(\xi_{k,j}/\xi_0)$, where $\xi_0=2^{15}$ is the initial state
and $\xi_{k,j}\in[2^{15},2^{31})$ is the terminal state of lane $j$ in
tile $k$. Summing over lanes gives
\begin{equation}
\label{eq:codelength-balance}
\begin{aligned}
 I_{\mathrm{state}}
 &=\sum_{k=1}^{n_{\mathrm{tile}}}\sum_{j=1}^{J}
 \log_2\!\left(\frac{\xi_{k,j}}{\xi_0}\right),\\
 L_{\mathrm{payload}}
 &=C_{\mathrm{CE}}-I_{\mathrm{state}}+\varepsilon_{\mathrm{rANS}}.
\end{aligned}
\end{equation}
Under the log-uniform approximation $\rho(\xi)\propto1/\xi$ for
terminal ANS states~\citep{duda_ans}, the log-state increase
$\log_2(\xi/\xi_0)$ is uniform on $[0,16)$, giving
\begin{equation}
\label{eq:codelength-state-mean}
 \mathbb{E}\!\left[\log_2(\xi/\xi_0)\right]
 \approx\frac{1}{16}\int_0^{16}u\,du=8,
 \qquad
 \mathbb{E}[I_{\mathrm{state}}]\approx8J n_{\mathrm{tile}}.
\end{equation}
During scale search, EntroPack computes $C_{\mathrm{CE}}$ from symbol counts
before the terminal states are known. Subtracting the upper bound
$16J n_{\mathrm{tile}}$ gives $C_{\mathrm{CE}}-16J n_{\mathrm{tile}}$,
a lower bound on the modeled payload term $C_{\mathrm{CE}}-I_{\mathrm{state}}$
in \eqref{eq:codelength-balance}. EntroPack omits $\varepsilon_{\mathrm{rANS}}$
and uses this lower bound as its estimate before rounding.
Clipping the estimate at zero and rounding up to a whole number of
16-bit payload words gives the predicted payload length in \eqref{eq:estimate},
\[
 \widehat L_{\mathrm{payload}}
 =16\left\lceil
 \frac{\max(0,C_{\mathrm{CE}}-16J n_{\mathrm{tile}})}{16}
 \right\rceil .
\]

\paragraph{Fixed-bitstream validation.}
We fix the fields, probability tables, tiling, and coding precision of the
$10240\times3840$ Z-Image-Turbo reference layer at an 8 bpp target.
With 12-bit frequencies and 10,803 tiles of at most 455 vectors, the
encoder emits $L_{\mathrm{payload}}=304{,}722{,}656$ bits, excluding
metadata. Table~\ref{tab:codelength-estimate} shows how the deduction per
lane affects the predicted payload.

Cross-entropy alone overestimates the payload by 0.91\%. The 16-bit
lower-bound deduction underestimates its length by
0.91\%, and the 8-bit mean underestimates it by 0.00027\%. Both corrected
estimates are within 1\% of the measured payload on this stream.
EntroPack uses the lower bound during scale search so that the state
approximation can only reduce the modeled payload length.

\begin{table}[!htb]
\centering
\footnotesize
\setlength{\tabcolsep}{4pt}
\renewcommand{\arraystretch}{1.08}
\begin{tabular}{cccc}
\toprule
Method & Deduction (bits/lane) & Payload (Mbit) & Estimate $-$ measured (bits) \\
\midrule
Measured rANS output & --- & 304.72 & --- \\
\midrule
Cross-entropy alone & 0 & 307.49 & $+2{,}764{,}745$ \\
EntroPack (Eq.~\ref{eq:estimate}) & 16 & 301.96 & $-2{,}766{,}384$ \\
Log-uniform mean & 8 & 304.72 & $-816$ \\
\bottomrule
\end{tabular}
\caption{Effect of the terminal-state deduction on the accuracy of rANS
payload length estimates.}
\label{tab:codelength-estimate}
\end{table}

\FloatBarrier
\subsection{Row Subsampling Ablation for Scale Selection}
\label{sec:subsampled-rate-search}

To reduce the cost of scale selection, EntroPack samples rows once and
uses those rows to estimate the rate at every candidate scale in
\eqref{eq:rootfind}. It then compresses the full matrix at the selected
scale. Table~\ref{tab:coding} shows that row subsampling reduces the
time to encode Z-Image-Turbo from 5.74~s to 1.76~s at an 8~bpp target,
a $3.26\times$ speedup. We now vary the fraction of rows sampled in
each of Z-Image-Turbo's 68 weight matrices of shape $10240\times3840$.
To isolate the effect of row sampling from the error in the payload
estimate of \eqref{eq:estimate}, we evaluate each candidate scale by
encoding the sampled rows with rANS.

\begin{figure}[!htb]
\centering
\includegraphics[width=0.98\linewidth]{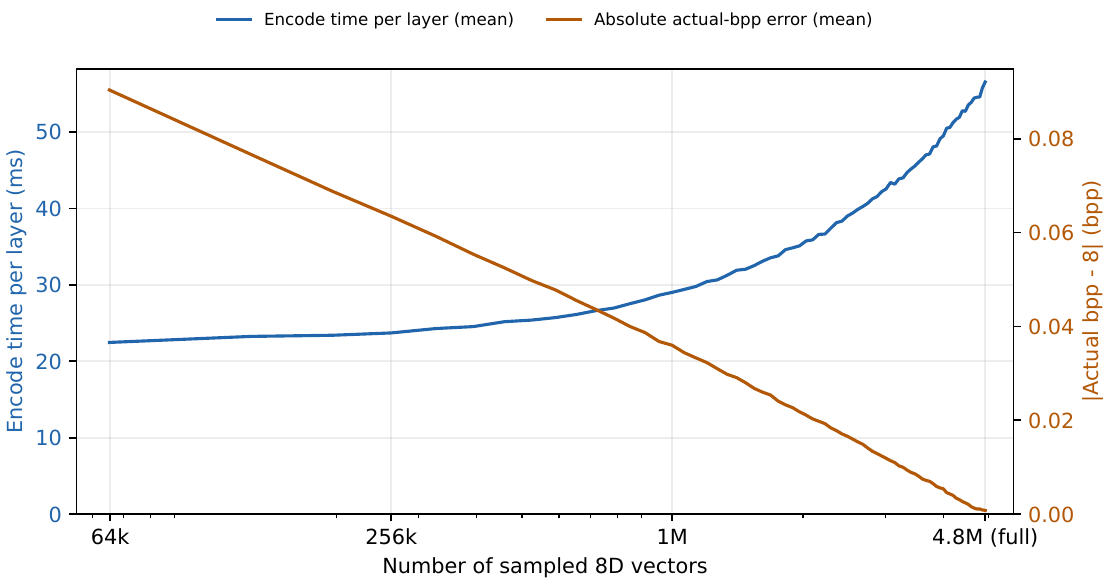}
\caption{Effect of sample size on encoding time and absolute rate error,
averaged over layers.}
\label{fig:subsampled-rate-search}
\end{figure}

Figure~\ref{fig:subsampled-rate-search} shows that the mean absolute
deviation from the 8~bpp target decreases steadily as the number of
sampled 8D vectors increases. Mean encoding time per layer changes little at small sample
sizes and rises as the search approaches full sampling. EntroPack
treats the sampling budget as a tunable hyperparameter,
with a default of 262,144 eight-dimensional vectors per layer.

\FloatBarrier
\subsection{Optional per-row rate--distortion refinement}
\label{sec:exp-rdo-net}

Per-row refinement seeks lower weight error through additional encoding
work. Candidate count controls the available row resolutions,
while sweep count controls updates to the row allocation and shared
probability model. We vary these settings separately on Z-Image-Turbo's linear
weights at a 3 bpp target. Allocations use the single-scale baseline's estimated
storage budget, and we compare reconstruction error and encoding time at
similar stored rates.

\begin{figure}[!htb]
\centering
\includegraphics[width=0.95\linewidth]{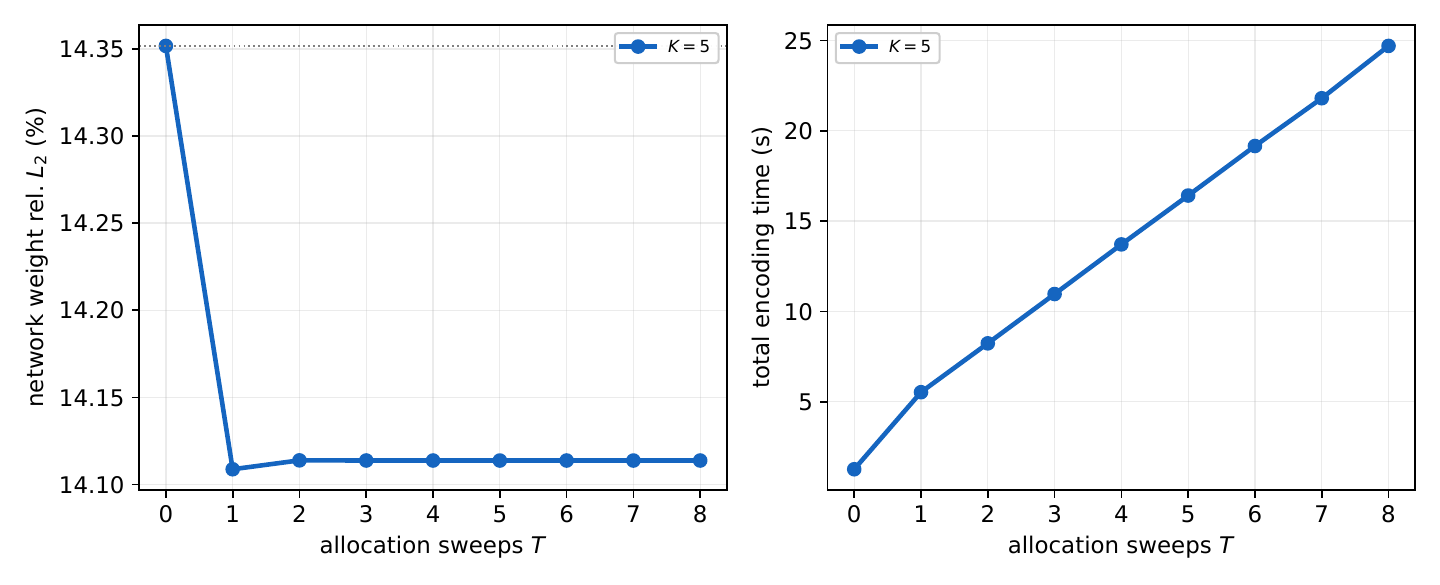}
\caption{RDO sweep count on Z-Image-Turbo at a target of 3 bpp with five candidate
scales. Left: network weight error. Right: total model encoding time.}
\label{fig:rdosweep}
\end{figure}

With five candidates per row, Figure~\ref{fig:rdosweep} varies the number
of allocation sweeps from the single-scale baseline at zero sweeps. Most
of the observed improvement occurs in the first sweep; further sweeps
increase encoding time with little change in reconstruction error.

\begin{figure}[!htb]
\centering
\includegraphics[width=0.95\linewidth]{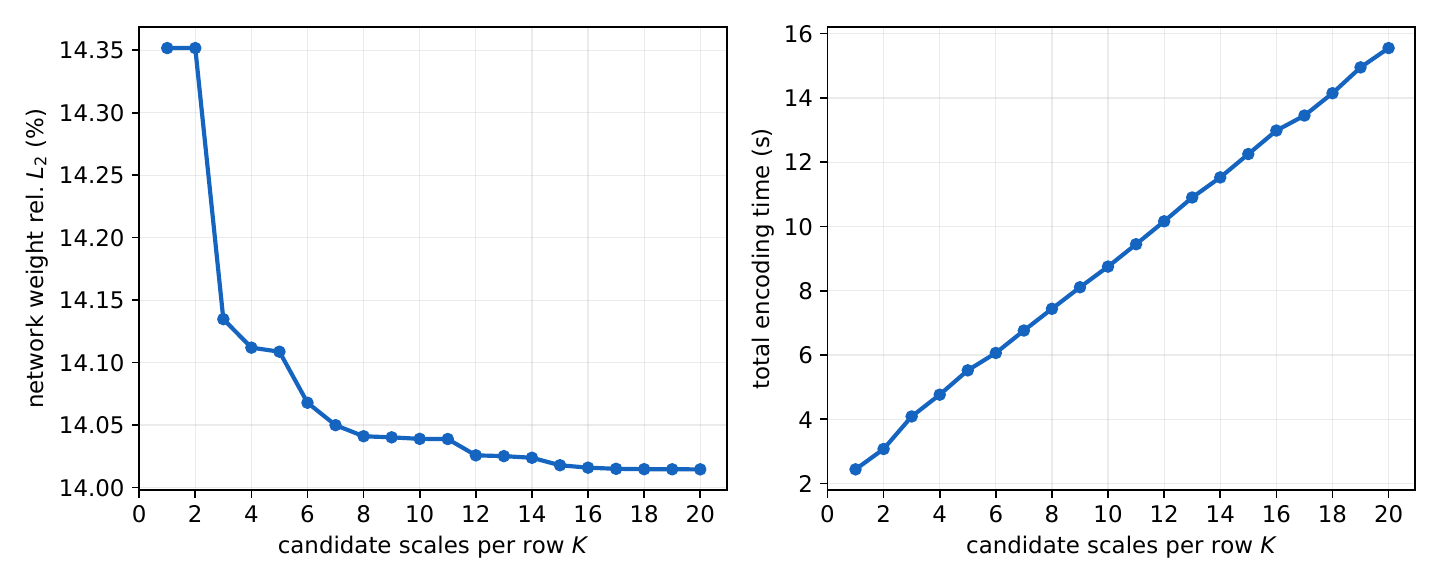}
\caption{RDO candidate count on Z-Image-Turbo at a target of 3 bpp with one allocation
sweep. Left: network weight error. Right: total encoding time.}
\label{fig:rdocands}
\end{figure}

We then hold the sweep count at one and vary the number of candidates in
Figure~\ref{fig:rdocands}. Candidate scale ratios form nested sets within
the fixed empirical range $[0.70,1.45]$ relative to the baseline scale.
Larger candidate sets reduce reconstruction error, with diminishing gains
as the set grows, while encoding time continues to increase.

\FloatBarrier
\subsection{Effect of Tile-Level Parallelism on Weight Decoding and Storage}
\label{sec:exp-cost}

\begin{figure}[!htb]
\centering
\includegraphics[width=0.95\linewidth]{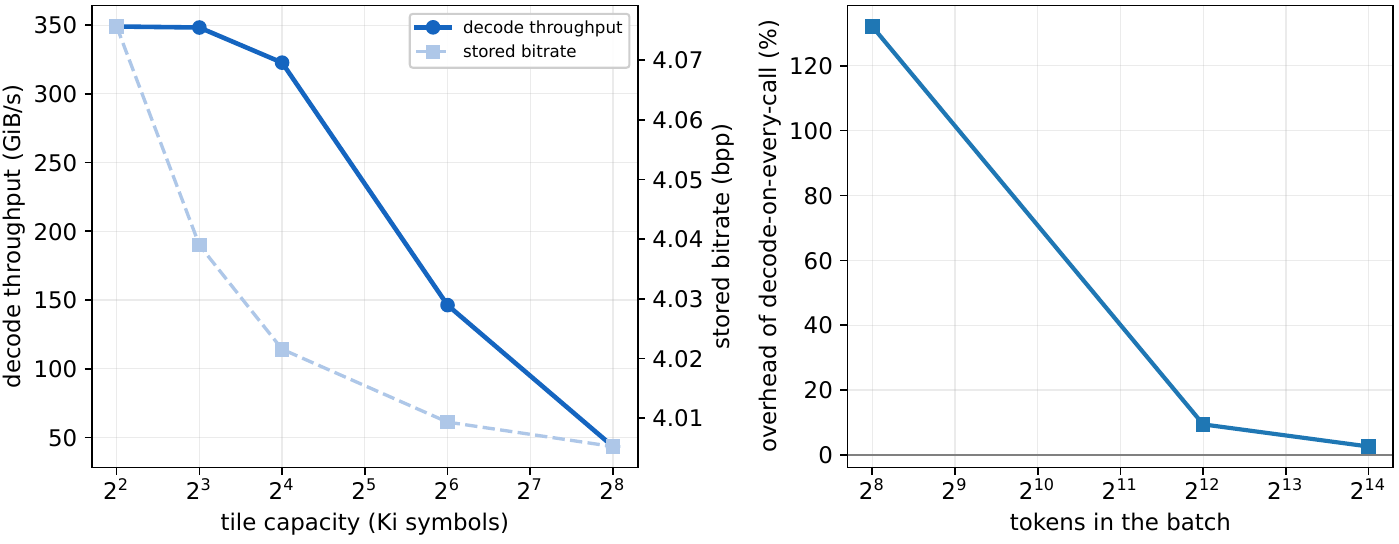}
\caption{Reconstruction cost on the reference layer at a 4 bpp target.
Left: decode throughput and stored rate versus tile size. Right: linear-layer
latency overhead relative to resident BF16 weights, versus the number of
input tokens processed together.}
\label{fig:cost}
\end{figure}

Inference overhead depends on decoder speed and the computation that
reuses each weight. Figure~\ref{fig:cost} uses the reference layer
compressed at a 4 bpp target.

The left panel of Figure~\ref{fig:cost} measures how tile size affects
decoding throughput and stored rate. Throughput counts reconstructed BF16
bytes per second, and stored rate includes all metadata. Over this sweep,
smaller tiles decode faster but use more storage for terminal states and
offsets, reflecting the cost of exposing more independent decoding work.

The right panel of Figure~\ref{fig:cost} holds the compressed weights fixed
and varies the number of input tokens processed in one linear-layer call.
For each input size, we compare a BF16 matrix product using resident,
uncompressed weights with a call that first reconstructs the weights and
then performs the same matrix product. The vertical axis reports the
percentage increase in total call time relative to the uncompressed case.
This overhead falls from 132\% at 256 tokens to 9.4\% at 4096 tokens and
2.6\% at 16,384 tokens. Larger matrix products amortize reconstruction
over more computation.

Table~\ref{tab:tileoverhead} quantifies the tile metadata cost under the
default configurations. Each tile stores terminal rANS states and a payload
offset so that decoding can start independently. The table reports this
tile metadata storage in bytes and as a fraction of total compressed size.
It occupies 1.80--3.60\% of storage across the displayed targets and is
included in every reported bitrate.

\begin{table}[!htb]
\centering
\footnotesize
\setlength{\tabcolsep}{3pt}
\begin{tabular}{ccccccc}
\toprule
target bpp & tile (symbols) & tile (vectors) & tiles & $b$ & tile metadata (bytes) & tile metadata (\%) \\
\midrule
2 & 32,768 & 3,640 & 1,351 & 11 & 178,336 & 1.80 \\
4 & 16,384 & 1,820 & 2,701 & 11 & 356,536 & 1.80 \\
7 & 8,192 & 910 & 5,402 & 11 & 713,068 & 2.06 \\
8 & 4,096 & 455 & 10,803 & 12 & 1,426,000 & 3.60 \\
\bottomrule
\end{tabular}
\caption{Tile configurations and metadata storage on the reference layer. Tile metadata comprises rANS states and payload offsets, with percentages relative to total stored bytes.}
\label{tab:tileoverhead}
\end{table}

\FloatBarrier
\subsection{Effect of Kernel Fusion on Weight Decoding}
\label{sec:gpu-fusion}

EntroPack combines rANS decoding, parity recovery, $E_8$ point
reconstruction, and row rescaling in one GPU kernel. In this section,
we examine how much decoding time this fusion saves. Using Z-Image-Turbo
weights compressed at a target rate of 4~bpp, we compare the time to
reconstruct them in BF16 with one fused kernel or four separate kernels,
keeping the rANS loop fixed. Reported times are averages over five runs.

Table~\ref{tab:gpu-fusion} shows that fusion cuts the mean time to
reconstruct a layer from 0.448 to 0.146~ms. Across the measured layers,
these savings reduce total decoding time from 122.8 to 39.9~ms, a
$3.08\times$ speedup. Fusion eliminates three kernel launches per layer
and lets each lane carry the decoded fields and lattice coordinates
through reconstruction in registers, avoiding the global-memory writes
and reads between stages.

\begin{table}[!htb]
\centering
\footnotesize
\setlength{\tabcolsep}{5pt}
\renewcommand{\arraystretch}{1.08}
\begin{tabular}{ccccc}
\toprule
Decoder & \shortstack{Decoding kernels\\per layer} &
\shortstack{Total decode\\time (ms)} & \shortstack{Mean per\\layer (ms)} &
\shortstack{Time per million\\weights ($\mu$s)} \\
\midrule
EntroPack (unfused) & 4 & 122.820 & 0.44825 & 20.217 \\
EntroPack (fused) & 1 & 39.869 & 0.14551 & 6.563 \\
\bottomrule
\end{tabular}
\caption{Effect of kernel fusion on weight decoding time in EntroPack.}
\label{tab:gpu-fusion}
\end{table}

\FloatBarrier
\subsection{Inference Trade-offs of Compressed Weights}
\label{sec:runtime-deployment}

EntroPack saves GPU memory by storing weights in compressed form, at the
cost of reconstructing them on each forward pass. We break down inference
time for Z-Image-Turbo and Qwen3.8-27B to quantify this overhead in
diffusion denoising and language model inference. When a model requires
CPU offloading, compression can also speed up inference by keeping more
weights on the GPU and reducing transfers from the CPU. We estimate this
benefit for MiniMax-H3 at different bitrates under a 24~GiB model-state
budget.

\paragraph{Z-Image-Turbo denoising time.}
We profile Z-Image-Turbo denoising for each method and break down the
total GPU time per step into weight reconstruction, matrix multiplications
in the linear layers (GEMM), attention, and other GPU operations.

At the 4~bpp target, matrix multiplication accounts for 63.9\% of
EntroPack's denoising step in Table~\ref{tab:zimage-step-breakdown},
attention for 11.6\%, and other operations for 17.5\%. Weight
reconstruction takes the remaining 7.1\% (39.20~ms per step).
This component explains most of the difference in step time across
EntroPack, HRTN, and HyperQuant. All three spend nearly the same time
on matrix multiplication and attention, but weight reconstruction takes
68.07~ms for HRTN and 539.61~ms for HyperQuant.

\begin{table}[!htb]
\centering
\footnotesize
\setlength{\tabcolsep}{3pt}
\renewcommand{\arraystretch}{1.06}
\begin{tabular}{cccccccc}
\toprule
& \multicolumn{2}{c}{Weight storage (bpp)} & \multicolumn{5}{c}{GPU time per denoising step (ms)} \\
\cmidrule(lr){2-3}\cmidrule(lr){4-8}
Method & Target & Actual & Recon. & GEMM & Attention & Other & Total \\
\midrule
BF16 & 16 & 16.0000 & 0.00 & 351.65 & 63.71 & 95.93 & 511.29 \\
\midrule
INT8 (torchao) & 8 & 8.0120 & 106.45 & 351.01 & 63.65 & 39.46 & 560.57 \\
FP8 (torchao) & 8 & 8.0096 & 71.55 & 351.69 & 63.72 & 95.90 & 582.85 \\
INT8 convrot (comfy-kitchen) & 8 & 8.0096 & 25.64 & 274.69 & 63.67 & 101.91 & 465.90 \\
FP8 (comfy-kitchen) & 8 & 8.0000 & 5.93 & 172.70 & 63.65 & 107.19 & 349.47 \\
EntroPack & 8 & 8.0604 & 49.71 & 351.59 & 63.69 & 96.05 & 561.04 \\
EntroPack & 7 & 7.0707 & 45.50 & 351.62 & 63.71 & 96.13 & 556.97 \\
EntroPack & 6 & 6.0555 & 42.42 & 351.53 & 63.69 & 96.09 & 553.72 \\
\midrule
NF4 (bitsandbytes) & 4 & 4.5000 & 13.79 & 351.96 & 63.72 & 94.42 & 523.88 \\
FP4 (bitsandbytes) & 4 & 4.5000 & 13.78 & 351.94 & 63.73 & 96.07 & 525.52 \\
NVFP4 (torchao) & 4 & 4.5000 & 297.21 & 351.27 & 63.64 & 95.73 & 807.85 \\
MXFP4 (torchao) & 4 & 4.2500 & 293.71 & 350.86 & 63.66 & 96.15 & 804.38 \\
HRTN & 4 & 4.0601 & 68.07 & 349.06 & 63.22 & 95.76 & 576.11 \\
HyperQuant & 4 & 4.1481 & 539.61 & 350.59 & 63.60 & 95.93 & 1049.72 \\
SDNQ & 4 & 4.5000 & 18.21 & 351.69 & 63.70 & 96.04 & 529.64 \\
EntroPack & 4 & 4.0237 & 39.20 & 351.77 & 63.72 & 96.23 & 550.92 \\
EntroPack & 3 & 3.0205 & 36.80 & 351.64 & 63.70 & 96.24 & 548.38 \\
\bottomrule
\end{tabular}
\caption{GPU time per Z-Image-Turbo denoising step by operation.}
\label{tab:zimage-step-breakdown}
\end{table}

\FloatBarrier
\paragraph{Language model inference across workloads.}
Table~\ref{tab:llm} shows a larger relative cost of compression during token
decoding than during prefill. To explain this difference, we profile weight
reconstruction within each forward pass of Qwen3.8-27B at 4~bpp. We vary the
batch size for single-token decoding, the number of supplied continuation tokens
in one backbone pass for multi-token prediction, and the prompt length for prefill.

\begin{figure}[!htb]
\centering
\includegraphics[width=\linewidth]{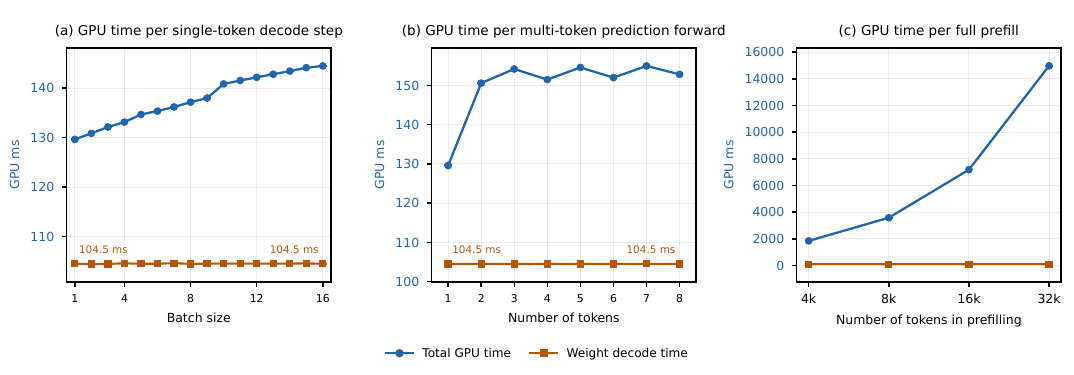}
\caption{Total GPU time and weight reconstruction time for Qwen3.8-27B at 4~bpp.
Panels (a) and (b) use an 8k-token prefix. Multi-token prediction and prefill
use batch size one. Curves show medians over five runs, with interquartile ranges shaded.}
\label{fig:llm-weight-scaling}
\end{figure}

Reconstructing the backbone weights takes about 104~ms, both in isolation and
across the workloads in Figure~\ref{fig:llm-weight-scaling}. In single-token
decoding, reconstruction accounts for 81\% of GPU time at batch size one and
72\% at batch size 16, where the sequences share the reconstructed weights.
Models that support multi-token prediction, such as Qwen3.8, can reuse the
reconstructed weights across several token positions in the same sequence.
Figure~\ref{fig:llm-weight-scaling}(b) shows that increasing the number of
positions processed together from one to eight reduces the share of GPU time
spent on weight reconstruction from 81\% to 68\%.
During prefill in panel~(c), the entire prompt shares one reconstruction,
reducing its share to just 0.7\% at 32k tokens.
This reuse makes on-demand weight reconstruction well suited to compute-intensive
workloads such as LLM prefill. Training passes that process many tokens together
could amortize the same cost. In small-batch autoregressive generation, weight
reconstruction lies on the critical path of each token, limiting EntroPack's use
in latency-sensitive serving. Reducing weight-decoding latency in this regime is
a priority for future work.

\FloatBarrier
\paragraph{MiniMax-H3 under a 24~GiB model-state budget.}
We model a 24~GiB GPU budget for the MiniMax-H3 diffusion transformer,
with a separate allowance for activations and temporary buffers.
For each format, we compare the time per denoising step with all weights
on the GPU and under the 24~GiB budget, where any CPU-resident weights
are loaded onto the GPU as needed during the step.

Under the budget in Table~\ref{tab:minimax-residency}, the CPU holds
37.72~GiB of the 61.72~GiB BF16 DiT. Loading these weights raises
denoising time from 26.25 to 32.19~s per step. At the 8~bpp target, only 6.88~GiB for INT8 (torchao)
and 7.19~GiB for EntroPack remain on the CPU, bringing step times down
to 27.94 and 27.74~s, respectively. At these similar step times,
EntroPack has about half the relative $L_2$ error of INT8
(0.53\% versus 1.04\%).

At the 4~bpp target, NF4 (bnb) and EntroPack keep the entire DiT on the GPU,
eliminating these transfers and yielding speedups of $1.23\times$ and
$1.22\times$ over BF16, respectively. The potential speedup is greater
when a larger fraction of the baseline weights resides on the CPU and
loading those weights accounts for a larger share of each denoising step.
EntroPack uses only 15.57~GiB at
this rate, leaving room to improve weight accuracy. Raising the target
to 6.12~bpp, the highest tested rate that fits, brings GPU model state
to 23.94~GiB and reduces relative $L_2$ error from 7.02\% to 1.70\%.
The denoiser still fits on the GPU, with 26.52~s per step compared with
26.44~s at 4~bpp.

\begin{table}[!htb]
\centering
\footnotesize
\setlength{\tabcolsep}{2pt}
\renewcommand{\arraystretch}{1.12}
\begin{tabular*}{\linewidth}{@{\extracolsep{\fill}}lccccccccc@{}}
\toprule
\multirow{2}{*}{Method} & \multicolumn{2}{c}{Weight storage (bpp)} &
\multicolumn{3}{c}{DiT state (GiB)} & Rel.\ $L_2$ &
\multicolumn{2}{c}{Time per denoising step (s)} & Speedup \\
\cmidrule(lr){2-3}\cmidrule(lr){4-6}\cmidrule(lr){8-9}
& Target & Actual & Total & GPU & CPU & (\%) &
All on GPU & 24~GiB budget & vs.\ BF16 \\
\midrule
NF4 (bnb) & 4    & 4.5000 & 17.39 & 17.39 & 0.00 & 9.14 & 26.26 & 26.26 & $1.23\times$ \\
EntroPack & 4    & 4.0330 & 15.57 & 15.57 & 0.00 & 7.02 & 26.44 & 26.44 & $1.22\times$ \\
EntroPack & 6.12 & 6.2045 & 23.94 & 23.94 & 0.00 & 1.70 & 26.52 & 26.52 & $1.21\times$ \\
EntroPack & 8    & 8.0889 & 31.19 & 24.00 & 7.19 & 0.53 & 26.59 & 27.74 & $1.16\times$ \\
INT8 (torchao) & 8 & 8.0000 & 30.88 & 24.00 & 6.88 & 1.04 & 26.85 & 27.94 & $1.15\times$ \\
BF16      & 16   & 16.0000 & 61.72 & 24.00 & 37.72 & 0.00 & 26.25 & 32.19 & $1.00\times$ \\
\bottomrule
\end{tabular*}
\caption{MiniMax-H3 DiT storage, weight reconstruction error, and denoising time under a 24~GiB model-state budget.}
\label{tab:minimax-residency}
\end{table}

\FloatBarrier
\subsection{Numerical dtype measurements}
\label{sec:exp-dtype-details}

EntroPack controls storage bitrate separately from the dtype of the
reconstructed weights. To evaluate this property, Table~\ref{tab:dtypes}
reports compression of seven dtype versions of the reference layer.
FP8 inputs use per-row scaling, and integer inputs use per-tensor scaling
before casting, with an offset for unsigned values. Each tensor is
reconstructed in its input dtype and compared with that input, so the
reported error measures compression after the initial format conversion.

Stored rates remain close to targets across these containers. Encoding
takes 5.1--8.4~ms and decoding 0.23--0.41 ms on this layer.

\begin{table}[!htb]
\centering
\footnotesize
\setlength{\tabcolsep}{3pt}
\begin{tabular}{cccccc}
\toprule
container & target bpp & stored bpp & rel.\ $L_2$ (\%) $\downarrow$ & encode (ms) $\downarrow$ & decode (ms) $\downarrow$ \\
\midrule
FP32 & 4.00 & 4.02 & 7.19 & 5.1 & 0.31 \\
FP32 & 8.00 & 8.05 & 0.51 & 8.2 & 0.41 \\
FP16 & 4.00 & 4.02 & 7.19 & 5.3 & 0.31 \\
FP16 & 8.00 & 8.05 & 0.51 & 8.4 & 0.40 \\
BF16 & 4.00 & 4.02 & 7.19 & 5.2 & 0.23 \\
BF16 & 8.00 & 8.05 & 0.53 & 8.4 & 0.34 \\
FP8 E4M3FN & 4.00 & 4.02 & 7.54 & 5.5 & 0.32 \\
FP8 E5M2 & 4.00 & 4.02 & 7.15 & 5.5 & 0.31 \\
INT8 & 4.00 & 4.02 & 8.17 & 5.1 & 0.32 \\
UINT8 & 4.00 & 4.02 & 7.76 & 5.1 & 0.32 \\
\bottomrule
\end{tabular}
\caption{Stored rate, reconstruction error, and codec time for seven dtypes on the reference layer. Errors use each dtype's input as the reference, with the UINT8 offset removed.}
\label{tab:dtypes}
\end{table}

\FloatBarrier
\section{Additional applications}
\label{sec:applications}

EntroPack can combine compressed storage with low-precision computation
and can assign different rates to individual layers. We evaluate these
two uses on Z-Image-Turbo, first retaining FP8 or INT8 matrix products and
then reallocating a fixed storage budget among layers.

\subsection{FP8 and INT8 computation with compressed storage}
\label{sec:exp-fp8route}

The dtype measurements above concern reconstruction of individual tensors.
Here we evaluate whether compressed FP8 and INT8 weights retain the runtime
benefits of low-precision computation during image generation. We first
convert Z-Image-Turbo's linear weights to row-scaled FP8 or INT8, then
compress them with EntroPack. During inference, weights are reconstructed
in those formats for low-precision matrix products. Table~\ref{tab:dividend}
compares these routes with uncoded FP8 and INT8, direct BF16 compression,
and the uncompressed BF16 model.

Both compressed low-precision routes reduce weight storage while preserving
faster denoising than the BF16 reference. At the 4 bpp target, they store
4.02 bpp, approximately half the 8.01 bpp of their uncoded counterparts.
Denoising steps take 392 and 393 ms, compared with 511 ms for BF16 and
343 ms for uncoded FP8. These times include weight reconstruction and
matrix multiplication.

Weight error uses the original BF16 weights as its reference. For FP8 and
INT8, it includes initial format conversion and compression. At the 6 bpp target,
direct BF16 compression has 1.89\% weight error, compared with 3.11\% for
the FP8 route and 2.37\% for the INT8 route.

\begin{table}[!htb]
\centering
\footnotesize
\setlength{\tabcolsep}{3pt}
\begin{tabular}{ccccc}
\toprule
route & stored bpp & rel.\ $L_2$ (\%) $\downarrow$ & PSNR $\uparrow$ & step (ms) $\downarrow$ \\
\midrule
BF16 reference & 16.00 & 0.00 & 99.00 & 511.3 \\
FP8 e4m3 container, uncoded & 8.01 & 2.65 & 22.21 & 343.1 \\
INT8 container, uncoded & 8.01 & 1.07 & 20.10 & 342.5 \\
EntroPack (4 bpp), FP8 container & 4.02 & 7.98 & 19.52 & 391.7 \\
EntroPack (4 bpp), INT8 container & 4.02 & 7.34 & 18.88 & 392.5 \\
EntroPack (4 bpp), direct BF16 & 4.02 & 7.18 & 19.62 & 550.9 \\
EntroPack (6 bpp), FP8 container & 6.06 & 3.11 & 22.31 & 392.2 \\
EntroPack (6 bpp), INT8 container & 6.06 & 2.37 & 20.12 & 392.0 \\
EntroPack (6 bpp), direct BF16 & 6.06 & 1.89 & 26.34 & 553.7 \\
\bottomrule
\end{tabular}
\caption{Storage, weight error, image fidelity, and runtime with compressed FP8 and INT8 weights on Z-Image-Turbo. Fidelity uses the uncompressed BF16 model as the reference.}
\label{tab:dividend}
\end{table}

\FloatBarrier
\subsection{Aggressive network-wide weight compression}
\label{sec:exp-mixed}

At very low bitrates, a uniform target may spend too little storage on
layers that strongly affect image structure. Table~\ref{tab:mixed} compares
uniform and mixed allocations of the same budget to Z-Image-Turbo's
linear weights. The mixed allocation assigns 3 bpp to modulation, embedding, and output layers and
approximately 2.47 bpp to the remaining layers. Both use approximately
2.5 stored bpp overall. At this matched storage, the mixed allocation
raises mean SSIM from 0.64 to 0.67.

\begin{table}[!htb]
\centering
\footnotesize
\setlength{\tabcolsep}{3pt}
\begin{tabular}{ccccc}
\toprule
allocation & network bpp & rel.\ $L_2$ (\%) $\downarrow$ & PSNR $\uparrow$ & SSIM $\uparrow$ \\
\midrule
uniform 2.5 bpp & 2.50 & 20.64 & 16.82 & 0.64 \\
sensitive@3 + rest@2.47 & 2.50 & 20.76 & 16.42 & 0.67 \\
\bottomrule
\end{tabular}
\caption{Uniform and mixed per-layer rate allocation on Z-Image-Turbo at approximately 2.5 stored bpp. Both allocations are evaluated against the same BF16 reference.}
\label{tab:mixed}
\end{table}

\begin{figure}[!htbp]
\centering
\begin{minipage}[t]{0.49\linewidth}
\centering
\includegraphics[width=\linewidth]{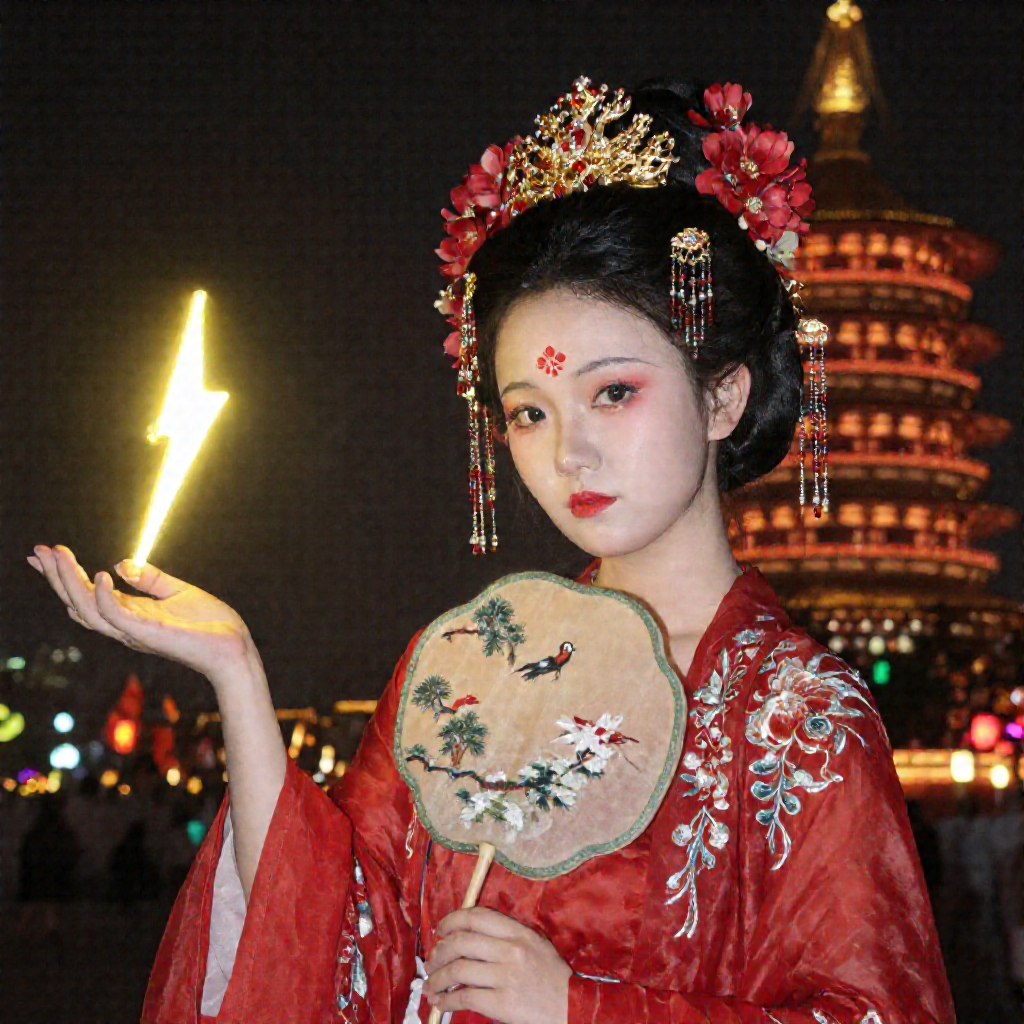}
\par\smallskip
{\small (a) Uniform allocation}
\end{minipage}\hfill
\begin{minipage}[t]{0.49\linewidth}
\centering
\includegraphics[width=\linewidth]{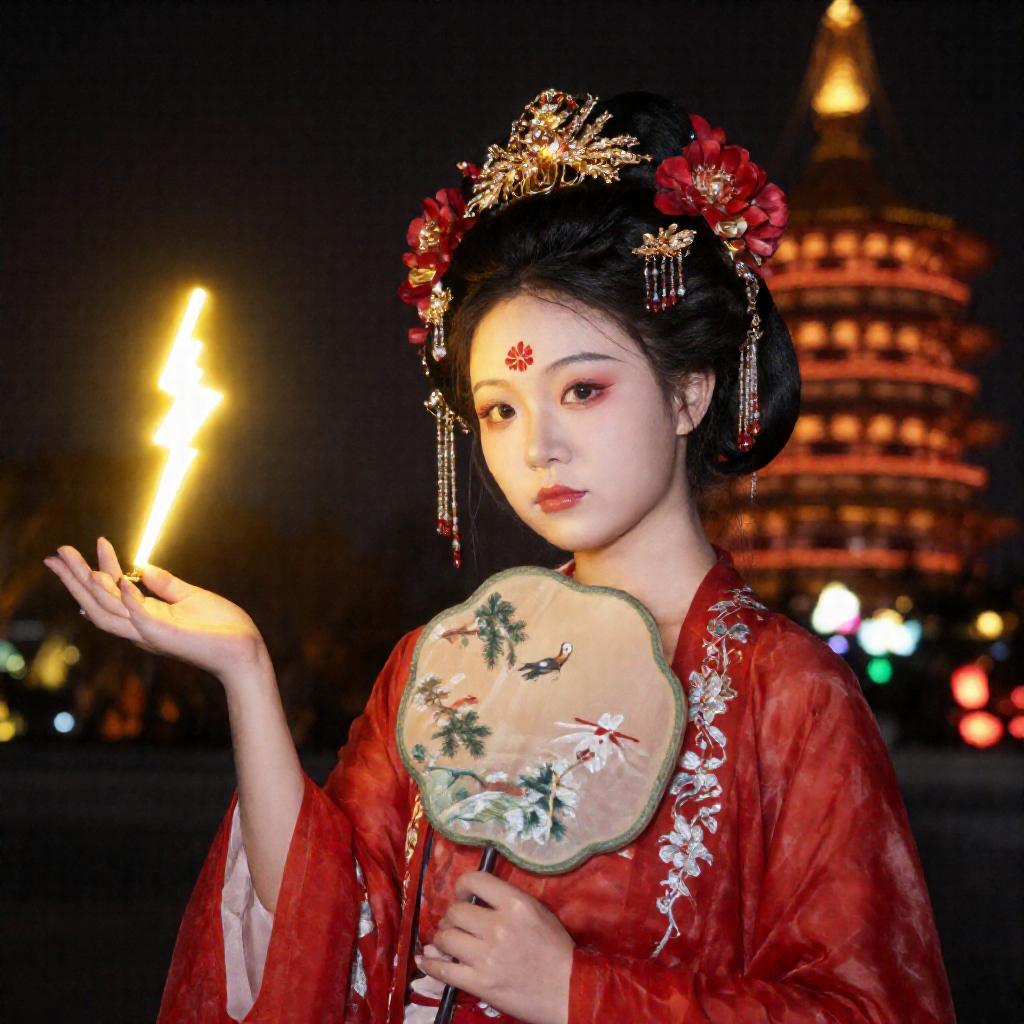}
\par\smallskip
{\small (b) Mixed allocation}
\end{minipage}
\caption{Z-Image-Turbo generations with uniform and mixed rate allocation at
2.50 stored bpp, using the same prompt, random seed, and denoising settings.}
\label{fig:mixed-images}
\end{figure}

Figure~\ref{fig:mixed-images} illustrates the effect of the allocation with
generations using the same prompt and seed. In this example, the mixed
allocation produces more consistent illumination on the face and raised
hand, improving the visual coherence of the generated image.

\FloatBarrier

\section{Additional methodological details}
\label{app:method-details}

We specify the search settings, per-row allocation algorithm, and GPU coding implementation.

\subsection{Scale-search settings}
\label{app:search}

The scale search in Algorithm~\ref{alg:rate} uses twelve
geometric-bisection steps on $[10^{-3},8]$ and a nominal sample budget of
$2^{18}$ eight-dimensional vectors. To compare candidate scales on the
same data, we sample complete rows with a shape-dependent seed, holding
them fixed throughout the search.

Probability-table precision $b$ determines the number of distinct symbols
that can receive positive frequencies. We start at 11 bits for targets
up to 7 bpp and at 12 bits above that. If a full-tensor field alphabet
exceeds $2^b$, we increase $b$ and repeat scale selection, up to 15 bits.
If the alphabet still exceeds this capacity, we coarsen the scale by setting
$s\leftarrow1.05\,s\,A_{\max}/2^b$ and requantizing until the alphabets fit,
where $A_{\max}$ is the largest coordinate-alphabet size.

\subsection{RDO allocation algorithm}
\label{app:rdo-algorithm}

Algorithm~\ref{alg:rdo} details the refinement in \S\ref{sec:rdo} under
the baseline's estimated storage budget. Candidate fields, fitted row
scales, and distortions are precomputed. Each sweep updates the shared probability model and selects row
candidates through a common rate-penalty search.

\begin{algorithm}[!htbp]
\caption{Per-row refinement using the baseline storage budget.}
\label{alg:rdo}
\begin{algorithmic}[1]
\Require Weights $W$, baseline $(s,b)$, ratios $\{\rho_k\}$ including $1$, sweeps $T$
\State For each $\rho_k$, quantize at $\rho_k s$ and fit row scales by \eqref{eq:scale_fit}
\State Record candidate fields, scales $\sigma_{r,k}$, and squared errors $D_{r,k}$
\State Initialize all $k(r)$ to the baseline; let $B$ be its estimated stored bytes
\For{$t=1,\ldots,T$}
  \State Fit the shared model $q_\theta$ to the current row selections
  \State $R_{r,k}\gets-\sum_{v\in r}\log_2q_\theta(c_v^{(k)},a_v^{(k)})$ for every $(r,k)$
  \State Estimate the current mixture's stored bytes $\widehat B_{\mathrm{mix}}$
  \State $R_{\mathrm{budget}}\gets\sum_r R_{r,k(r)}+8(B-\widehat B_{\mathrm{mix}})$
  \State Define $\kappa_\lambda(r)=\arg\min_k[D_{r,k}+\lambda R_{r,k}]$,\quad $L(\lambda)=\sum_r R_{r,\kappa_\lambda(r)}$
  \State $\widehat k\gets\kappa_0$
  \If{$L(0)>R_{\mathrm{budget}}$}
    \State Set $\lambda_{\mathrm{lo}}=0$ and bracket $\lambda$ by doubling a positive upper bound
    \For{$j=1,\ldots,20$}
      \State $\lambda\gets\lambda_{\mathrm{hi}}/2$ if $\lambda_{\mathrm{lo}}=0$, otherwise $\sqrt{\lambda_{\mathrm{lo}}\lambda_{\mathrm{hi}}}$
      \State Raise $\lambda_{\mathrm{lo}}$ to $\lambda$ if $L(\lambda)>R_{\mathrm{budget}}$; otherwise lower $\lambda_{\mathrm{hi}}$ to $\lambda$ and retain $\widehat k\gets\kappa_\lambda$
    \EndFor
  \EndIf
  \State $k\gets\widehat k$
\EndFor
\State \Return the selected fields and row scales $\{\sigma_{r,k(r)}\}$
\end{algorithmic}
\end{algorithm}

\FloatBarrier
\subsection{Tiled RANS implementation}
\label{app:tilerans}

\paragraph{Tile layout and storage.}
Independent tile decoding requires a saved set of rANS states and a
location in the payload. Each tile has $J=32$ 32-bit terminal states and
a 32-bit payload offset, requiring $4J+4=132$ bytes of tile metadata. For a configured
size of $S$ symbols, a tile holds $V=\max(32,\lfloor S/9\rfloor)$
eight-dimensional vectors. Table~\ref{tab:tileoverhead} lists the defaults,
from $S=32768$ at 2 bpp to $S=4096$ at 8 bpp. The tensor also stores a
final boundary offset and FP32 row scales.

\paragraph{State normalization.}
The implementation uses 32-bit rANS states maintained in
$[2^{15},2^{31})$ and 16-bit payload words. Renormalization transfers words
between the state and payload to preserve this range. The supported
probability precisions $b\in\{9,\ldots,15\}$ give table frequencies
$M=2^b\le2^{15}$, so at most one payload word is needed per symbol.

\paragraph{Decoder lookup tables.}
Lookup tables recover a symbol and its frequency interval from the low
bits of the current rANS state. They reside in shared memory or packed global memory,
depending on alphabet size and available device resources.

\section{Limitations and future work}
\label{sec:limitations}

EntroPack's flexible storage rates require additional work during model
conversion and inference. Scale search, probability modeling, and entropy
coding increase conversion cost relative to a single fixed-width
quantization pass. Reducing repeated data passes could lower this cost.
During inference, the decoder materializes a dense weight tensor before
each matrix product, making reconstruction harder to amortize in small
workloads. Fusing entropy decoding and weight reconstruction with matrix
multiplication could avoid this intermediate tensor and reduce memory traffic.

Activations and key--value caches are produced during inference and would
require online compression. Extending EntroPack beyond static weights to
these tensors would therefore require probability modeling and rate
selection to track changing distributions with low latency.

\end{document}